\documentclass{article}

\usepackage[preprint]{neurips_2026}

\usepackage[utf8]{inputenc}
\usepackage[T1]{fontenc}
\usepackage{hyperref}
\usepackage{url}
\usepackage{booktabs}
\usepackage{amsfonts}
\usepackage{amsmath}
\usepackage{amssymb}
\usepackage{nicefrac}
\usepackage{microtype}
\usepackage{xcolor}
\usepackage{graphicx}
\usepackage{float}
\usepackage{multirow}
\usepackage{flafter}  
\usepackage{placeins}

\graphicspath{{figures/}}

\hypersetup{
  pdftitle={Meaning Contracts: Local Semantic Geometry in Large Language Models},
  pdfauthor={Andr\'es Daniel Godoy Ortiz}
}

\title{Meaning Contracts: Local Semantic Geometry in Large Language Models}

\author{%
  Andr\'es Daniel Godoy Ortiz \\
  Imperial College London \\
  \texttt{ddg225@ic.ac.uk}
}

\begin{document}

\maketitle

\begin{abstract}
This paper studies the local geometry of semantic meaning in large language models. Current representation analyses ask whether meanings are close in embedding space, or whether linear probes separate semantic classes, but they do not characterize the within-concept local geometry: what is invariant across realizations of a single semantic factor, what varies, and what changes when meaning is refined. This paper characterizes that geometry through two controlled probes evaluated on four architecturally diverse decoder-only LLMs (Gemma 4, Llama 3.2, Qwen 2.5, Pythia). A cross-lingual probe uses language as a perturbation to separate exact meaning matches, same-factor realizations, nearby concepts, and out-of-domain contrasts. A refinement probe separates specificity-bearing semantic narrowing from high-probability attributive elaboration and adversarial controls; both probes are validated by blind human judgments. Four findings recur across all models: once dominant variance directions are removed, sentence pairs order by semantic relatedness (same factor, then same domain, then different domain) without overlap over a band of layers; semantic factors organize around an invariant core surrounded by structured realization-specific variation; refinement produces directed displacement into tighter neighborhoods, including against controls built on the same base noun; and sentence-level versus token-local readouts expose complementary geometry---volume contraction at the sentence level, sharper anisotropic endpoints at token-local readout. We interpret these results as evidence that LLM semantics are organized as local anisotropic neighborhoods in which meaning contracts under refinement.
\end{abstract}

\section{Introduction}

A growing body of work has established that LLM hidden states encode semantically meaningful structure. Linear probes recover syntactic relations \citep{HewittManning2019}, spatial and temporal coordinates \citep{GurneeTegmark2024}, truth values \citep{MarksTegmark2024,LiEtAl2023ITI}, and factual associations \citep{HernandezEtAl2024}. The linear representation hypothesis formalizes the intuition behind these results: high-level concepts are represented as directions in activation space \citep{ParkEtAl2024,Mikolov2013,ElhageEtAl2022}. This hypothesis now underpins practical methods---representation steering \citep{ZouEtAl2023,TurnerEtAl2023}, latent-space reward shaping, and safety-relevant anomaly detection---that all intervene on the geometric structure of hidden states.

Yet there is a gap between knowing that concepts \emph{have} a direction and knowing what the \emph{local neighborhood} around that direction looks like. Existing analyses treat semantic geometry globally: they ask whether two meanings are close, whether a probe separates classes, or whether a steering vector moves output in the intended direction. They do not characterize the local shape of the region that a semantic factor occupies: whether the factor has an invariant core or only a diffuse cloud, whether different realizations of the same meaning spread along structured directions or scatter isotropically, or what geometric signature distinguishes genuine semantic narrowing from superficial lexical elaboration. Without answers to these questions, cross-lingual alignment \citep{Cao2020,PengSogaard2024,WuEtAl2024,Conneau2020} tells us that representations \emph{coincide} but not how semantic neighborhoods are \emph{shaped}; representational change under refinement could reflect stronger meaning, lexical accumulation, or nuisance geometry; and steering interventions operate on directions whose local structure remains uncharacterized.

This paper addresses the gap with two controlled sentence-level probes evaluated on four decoder-only LLM families that differ materially in tokenizer, depth, width, attention configuration, and pretraining mixture: Pythia 2.8B \citep{BidermanEtAl2023}, Llama 3.2 3B \citep{GrattafioriEtAl2024}, Qwen 2.5 3B \citep{YangEtAl2024Qwen}, and Gemma 4 E4B \citep{GemmaTeam2026}. A \textbf{cross-lingual semantic-neighborhood probe} uses language as a controlled perturbation to separate exact meaning matches from same-factor realizations, nearby concepts, and out-of-domain contrasts. A \textbf{semantic refinement probe} distinguishes genuine semantic narrowing (e.g., \emph{report} $\rightarrow$ \emph{audit report} $\rightarrow$ \emph{environmental audit report}) from matched controls that add compatible description to the same noun (\emph{report} $\rightarrow$ \emph{written report} $\rightarrow$ \emph{written dated report}), stack near-synonyms, or hedge scope. Cross-family replication is central to the claims.

\paragraph{The findings in plain terms.}
Stated without technical vocabulary, the paper reports four observations. (1) Inside a model, sentences that mean the same thing are closer to each other than to sentences about related or unrelated topics, but this only becomes visible after removing a few dominant directions that mostly encode things like language identity. (2) Two literal translations of one sentence sit very close together, while different ways of expressing the same idea spread out around them along consistent directions. (3) When a phrase is made more specific (\emph{report} $\rightarrow$ \emph{environmental audit report}), its representation moves farther from the starting point and its different wordings cluster more tightly than when the phrase is merely spelled out (\emph{report} $\rightarrow$ \emph{written dated report}). (4) Averaging over a sentence and reading only its final token show two different aspects of this tightening.

The contributions are four empirical findings that hold across all four model families, together with the two probe datasets\footnote{Probe datasets, analysis code and the outputs behind every figure and table: \url{https://github.com/adgodoyo/Meaning_Contracts}.}:
\begin{itemize}
    \item \textbf{Depth-local semantic order.} Over a band of layers in each network, hidden states contain a semantic ordering that becomes visible once dominant high-variance directions are removed via PCA deflation \citep{MuViswanath2018}, with behavioral consequences for cross-lingual retrieval. ``Depth'' refers throughout to layer depth in the network, not to the number of removed components.
    \item \textbf{Invariant core with structured variation.} Exact meaning matches are always the nearest category under covariance-weighted distance---relative to the ambient variance of the layer---surrounded by structured, realization-specific local variation.
    \item \textbf{Refinement as directed contraction.} Semantic refinement is expressed as directed displacement into tighter local neighborhoods, producing more geometric change than matched controls, including controls built on the same base noun.
    \item \textbf{Layered readout geometry.} Sentence-level and token-local readouts expose complementary aspects: neighborhood contraction at the sentence level, sharper anisotropic endpoints at token-local readout.
\end{itemize}

These findings support a view of LLM semantics as \emph{local anisotropic neighborhoods}. We call the C1 region a factor's \emph{invariant core}, and the refinement-specific decrease in log-volume \emph{contraction}. This complements the linear representation hypothesis: where that hypothesis identifies directions, the present work describes the neighborhoods around them. A possible downstream implication, untested here, is that methods such as steering vectors \citep{TurnerEtAl2023,ZouEtAl2023}, latent-space interventions \citep{LiEtAl2023ITI}, and representation monitoring \citep{BurnsEtAl2023} could exploit not just direction but also local spread, anisotropy, and contraction as diagnostic or control signals.

\section{Related Work}

\paragraph{Probing and structural analysis.}
Structural probing \citep{HewittManning2019} and the broader probing methodology \citep{Belinkov2022,VoitaTitov2020,Tenney2019} have shown that LLM hidden states encode recoverable syntactic and semantic structure. These analyses establish \emph{that} structure exists; the present work characterizes \emph{what shape} it takes locally. The distinction matters because probing accuracy does not reveal whether a concept occupies a point, an isotropic cloud, or an anisotropic neighborhood.

\paragraph{Anisotropy and representation geometry.}
Contextual representations concentrate in a narrow cone of the ambient space \citep{Ethayarajh2019,MuViswanath2018,GaoEtAl2019}, and subsequent work has studied how this anisotropy varies across layers and models \citep{CaiEtAl2021,RajaeePilehvar2022,BisEtAl2021}. The ``all-but-the-top'' post-processing of \citet{MuViswanath2018} removes dominant directions to improve isotropy. The present work builds on this by showing that deflation does not merely improve downstream metrics but reveals a latent tiered ordering of semantic categories that is otherwise obscured by dominant variance directions.

\paragraph{Linear representations and concept geometry.}
The linear representation hypothesis posits that high-level concepts correspond to directions in activation space \citep{ParkEtAl2024,Mikolov2013,ElhageEtAl2022}. Empirical support includes linear probes for truth \citep{MarksTegmark2024,BurnsEtAl2023}, space and time \citep{GurneeTegmark2024}, and factual relations \citep{HernandezEtAl2024}. Representation engineering \citep{ZouEtAl2023} and activation addition \citep{TurnerEtAl2023} exploit this linearity for model steering, while inference-time intervention \citep{LiEtAl2023ITI} shifts activations along truthfulness-correlated directions. Closest to our question, \citet{ParkEtAl2025} show that hierarchical relations such as hypernymy are encoded as orthogonality between concept vectors. Their unit of analysis is a word-level concept estimated in unembedding space, and neither that formalism nor the linear representation hypothesis contains a term for variation \emph{within} a concept, because neither has an object corresponding to two different expressions of the same meaning. That object is our unit of analysis: a set of controlled sentence-level realizations of one fixed semantic factor, whose centroid displacement, effective dimensionality and log-volume we track across the full hidden-state stack. We are not aware of prior work that separates specificity-bearing narrowing from matched attributive elaboration and measures local dispersion, across depth, under that contrast.

\paragraph{Cross-lingual alignment and compositional change.}
Multilingual models develop partially aligned representations across languages \citep{Cao2020,Conneau2020,PengSogaard2024,WuEtAl2024,ChiEtAl2020}, typically measured by retrieval accuracy or classifier transfer. The cross-lingual probe introduced here goes further by decomposing cross-lingual similarity into an invariant core and structured realization-specific variation. Prior work on compositional representations \citep{Andreas2019,MitchellLapata2010} has studied how meaning composes from parts; the refinement probe provides a controlled test of a specific geometric prediction---that narrowing meaning should produce directed displacement \emph{and} neighborhood contraction, separable from generic elaboration.

\section{Method}
\subsection{Datasets and hidden-state extraction}

The first dataset is a \textbf{cross-lingual semantic-neighborhood probe} containing $1{,}200$ sentences built from $40$ semantic factors across $10$ domains and $5$ languages (English, German, Spanish, Arabic, Chinese). Each factor-language record contains $6$ sentences: $2$ exact parallel realizations and $4$ independent realizations, for $200$ factor-language records. A \emph{factor} is a specific semantic property, such as \emph{hot}. A nearby factor within the same domain might be \emph{warm} or \emph{cold}. Independent realizations need not be translations; for example, ``the object is hot'' and ``It has a very high temperature'' preserve the same factor while changing realization. This construction enables comparison of exact meaning matches, same-factor variation, same-domain nearby concepts, and clearly out-of-domain contrasts. Language is used as a control device, not as the endpoint of analysis.

Seven pair categories are derived from this probe: exact meaning match across languages (C1); same factor in the same language (C2); same factor across different languages (C3); same domain but different factor in the same language (C4); same domain but different factor across languages (C5); different domain in the same language (C6); and different domain across languages (C7).

The second dataset is a \textbf{semantic refinement probe} containing $3{,}840$ sentences in $48$ semantic chains. Each chain progresses across five refinement levels, denoted $S_0$ through $S_4$, from a coarse starting concept ($S_0$) to a progressively more specific one ($S_4$). The refinement arm adds specificity-bearing content (e.g., $S_0{=}$\emph{report} $\rightarrow$ $S_1{=}$\emph{audit report} $\rightarrow$ $S_2{=}$\emph{environmental audit report}); the main control arm adds high-probability attributive content that stays within the typical distribution of the base concept ($S_0{=}$\emph{report} $\rightarrow$ $S_1{=}$\emph{written report} $\rightarrow$ $S_2{=}$\emph{written dated report}), without displacing the representation toward a marked semantic subtype; adversarial controls stack near-synonyms or hedge scope. The $48$ chains ($20$ refinement, $20$ control, $8$ adversarial) are each realized with $4$ template families and $4$ surface variants at each of the $5$ levels ($80$ sentences per chain). Two registers, \emph{everyday noun} and \emph{formal compound}, provide stylistic variation across chains rather than defining extra refinement levels. Seven base nouns (\emph{certificate, form, invoice, license, notice, record, report}) appear in both arms, which allows a same-base comparison (Section~\ref{sec:refinement_results}). Token length is controlled for in Appendix~\ref{sec:token_length}.

\paragraph{Semantic refinement, defined.}
A step $S_{k}\rightarrow S_{k+1}$ counts as \emph{refinement} when (i) the head category is preserved; (ii) the more specific expression ordinarily entails its predecessor; (iii) the reverse inference does not hold; and (iv) the step introduces a restriction that moves the concept toward a marked semantic subtype, rather than explicitly stating a high-probability default attribute of the base concept. Condition (iv) separates the arms: \emph{environmental audit report} commits \emph{report} to a lexicalised subtype, whereas \emph{written dated report} makes explicit what is already typical of \emph{report}. Control steps therefore also narrow the extension of the base noun in a weak sense; the refinement-minus-control contrast should accordingly be read as a conservative lower bound on the effect of marked narrowing, not as a contrast against zero narrowing.

\paragraph{Construction.}
Both probes were specified semantically before any sentence was written, and the specification was frozen before realization. The cross-lingual probe contains no model-generated text: its $40$ factors were hand-authored, each with an explicit canonical meaning and six realizations typed as close parallels or independent expressions of the same property, which is what makes the seven pair categories constructible from metadata rather than assigned per pair. English realizations were written against the specification and every sentence in the other four languages was reviewed and revised by a native speaker. For the refinement probe, every factor, chain, control and arm definition was specified by hand, together with seed sentences; each chain narrows along a dimension declared in advance (e.g., setting, material, population, technology, size), recorded in the released data beside a gloss of the intended meaning. Surface strings were then instantiated under the frozen schema with model assistance and checked against the specification, and token length across levels was equalized manually, sentence by sentence. No model assigned any label in either probe. Holding token length near-constant required accepting some awkward phrasings; we took that trade-off deliberately, since length differences would contaminate every distance measured, and test its consequences in Appendix~\ref{app:qc}. Three blind human-validation studies of both probes are reported in Appendix~\ref{app:human}.

\begin{table}[t]
\centering
\caption{How the two probes are constructed.}
\label{tab:datasets}
\begin{tabular}{llr}
\toprule
Probe & Construction & Total \\
\midrule
Semantic-neighborhood & $10$ domains $\times$ $4$ factors $\times$ $5$ languages $\times$ $6$ sentences & $1{,}200$ \\
Semantic refinement & $48$ chains $\times$ $5$ levels $\times$ $4$ templates $\times$ $4$ variants & $3{,}840$ \\
\bottomrule
\end{tabular}
\end{table}

All analyses are training-free. Hidden states are extracted from frozen LLMs and controlled sentence groups are compared across layers, readouts, nuisance-removal conditions, and model families. Pythia 2.8B \citep{BidermanEtAl2023}, Llama 3.2 3B \citep{GrattafioriEtAl2024}, Qwen 2.5 3B \citep{YangEtAl2024Qwen}, and Gemma 4 E4B \citep{GemmaTeam2026} serve as cross-family controls: these differ in tokenizer, pretraining mixture, and architecture. For each model and sentence, the full hidden-state stack is extracted (Table~\ref{tab:models}). Two readouts are analyzed: \textbf{mean pooling} (average over non-padding tokens) as the sentence-level readout, and \textbf{last-token pooling} as the token-local readout. Keeping them separate is important: later results show they expose different parts of the same geometry.

\begin{table}[t]
\centering
\caption{Models analyzed. Layer counts include the input embedding state.}
\label{tab:models}
\begin{tabular}{lccc}
\toprule
Model & Family & Hidden-state stack & Width \\
\midrule
Pythia 2.8B & GPT-NeoX style decoder & $33$ & $2560$ \\
Llama 3.2 3B & Llama-family decoder & $29$ & $3072$ \\
Qwen 2.5 3B & Qwen-family decoder & $37$ & $2048$ \\
Gemma 4 E4B & Gemma-family decoder & $43$ & $2560$ \\
\bottomrule
\end{tabular}
\end{table}

\subsection{Recovering semantic order}
\label{sec:order_method}

The working assumption is that raw hidden states mix semantic structure with dominant high-variance directions \citep{Ethayarajh2019,MuViswanath2018,GaoEtAl2019}. Raw and top-$k$-deflated representations are therefore compared throughout.

\paragraph{Top-$k$ PCA deflation.}
Given sentence representations $X \in \mathbb{R}^{N \times D}$ at a fixed layer, centering and removing the projection onto the top $k$ right singular vectors $V_k$ yields: $X^{(-k)} = \tilde{X} - (\tilde{X}V_k^\top)V_k$. This is the contextual analogue of ``all-but-the-top'' \citep{MuViswanath2018}. A sweep over $k=1,\dots,10$ is performed.

\paragraph{Distances and aggregation.}
Three metrics are used: cosine similarity, Euclidean distance, and truncated Mahalanobis distance. Let $\Sigma_\ell \approx U_r \Lambda_r U_r^\top$ be the eigendecomposition of the sentence-state covariance at layer $\ell$. The truncated Mahalanobis distance is
\begin{equation}
d_M(x,y)^2 = (x-y)^\top U_r \Lambda_r^{-1} U_r^\top (x-y),
\end{equation}
i.e., Euclidean distance after whitening by the leading covariance eigenspectrum. $\Sigma_\ell$ is estimated once per layer from all $1{,}200$ probe sentences pooled, without access to pair-category labels, and is shared by all categories. Truncation is necessary because the empirical covariance is estimated in a high-dimensional regime, and inverting the full spectrum would over-weight unstable tail directions. The metric is therefore restricted to the best-estimated leading eigenspace, using $r=128$ for mean-pooled and $r=256$ for last-token states; Appendix~\ref{app:rank} shows that both values lie on a plateau of the rank sweep, except Gemma under last-token readout, where the plateau begins at $r=512$, and Appendix~\ref{app:indepcov} that the ordering survives covariance estimated on disjoint sentences. Category-level values are means over all pairs in a category; reweighting so that every factor, domain or language contributes equally changes category means by less than $0.1\%$.

\paragraph{Two orderings.}
To compare metrics on one scale, define closeness $\kappa_m(x,y)$ as cosine
similarity for cosine and as $-d_m(x,y)$ for Euclidean and Mahalanobis
distance, so that larger always means closer. Let $\bar\kappa_c(\ell,k)$ be the
mean closeness over the pairs in category $c$ at layer $\ell$ and deflation
level $k$. We define two orderings once and test every metric and readout
against both, at each layer independently.

\emph{Relatedness ordering (R).} Pairs are closer the more semantically related
they are. With $G_1=\{C1,C2,C3\}$ (same factor), $G_2=\{C4,C5\}$ (same domain)
and $G_3=\{C6,C7\}$ (different domain), R holds at $(\ell,k)$ when
$\min_{c\in G_1}\bar\kappa_c > \max_{c\in G_2}\bar\kappa_c$ and
$\min_{c\in G_2}\bar\kappa_c > \max_{c\in G_3}\bar\kappa_c$; crossings within a
group are allowed.

\emph{Core-and-spread ordering (B).} Exact matches are closest, freely realized
same-factor pairs farthest, and all inter-concept pairs lie between. With
$M=\{C4,\dots,C7\}$, B holds when
$\bar\kappa_{C1} > \max_{c\in M}\bar\kappa_c$ and
$\min_{c\in M}\bar\kappa_c > \max(\bar\kappa_{C2},\bar\kappa_{C3})$.
When, in addition, $\min(\bar\kappa_{C4},\bar\kappa_{C5}) >
\max(\bar\kappa_{C6},\bar\kappa_{C7})$, we say B holds with an ordered middle
band, written B$^{+}$. The letter refers to the three bands---core, middle band and outer realizations---and B is the ordering tested in the rank and covariance checks of Appendices~\ref{app:rank} and~\ref{app:indepcov}.

\paragraph{Reference frames.}
The two orderings correspond to two ways of measuring displacement. Cosine
similarity measures it in absolute units, so the high-variance
directions along which concepts differ dominate, and the natural test is R.
Mahalanobis distance measures it relative to the layer's ambient variance:
each eigendirection is rescaled by $1/\sqrt{\lambda_i}$, which compresses the
high-variance directions separating concepts and stretches the low-variance
directions along which realizations of a single factor differ. In this frame
inter-concept pairs compress into a middle band and realization spread is
magnified, which is what B describes. Euclidean distance on raw states is also absolute, but once the leading directions are removed it is no longer tied to one frame, and where it lands depends on the readout (Section~\ref{sec:order_results}). Table~\ref{tab:orderings} reports, for every metric and readout, the share of layers
satisfying each ordering, in raw space and at the deflation level used in the figures.

\paragraph{Behavioral validation.}
Cross-lingual retrieval is evaluated by nearest-neighbor search under cosine similarity in raw and deflated spaces. Each query sentence in a source language must retrieve its factor-matched sentence from a pool of 40 candidates in the target language, giving a chance Recall@1 of $1/40 = 0.025$.

\subsection{Measuring refinement geometry}

Chain-level summaries aggregate sentence realizations per chain, layer, and level. Let $c \in \{1,\dots,48\}$ index chains and $\ell$ index layers. Directed displacement is measured from the base level to later levels:
\begin{equation}
\Delta_{c,\ell}^{(0 \rightarrow k)} = d\!\left(h_{c,\ell}^{(S_0)},
h_{c,\ell}^{(S_k)}\right), \qquad k \in \{1,\dots,4\},
\end{equation}
where $h_{c,\ell}^{(S_k)}$ denotes the mean hidden state of
chain $c$ at level $S_k$ and layer $\ell$. The refinement-minus-control
difference in displacement measures the extra trajectory associated with
semantic narrowing.

\paragraph{Local spread measures.}
The covariance of the sentence cloud at each chain-level-layer is estimated from its $n=16$ realizations ($4$ templates $\times$ $4$ surface variants; all $240$ chain-level groups contain exactly $16$) and summarized via participation ratio $\mathrm{PR}(\Sigma) = (\sum_i \lambda_i)^2 / \sum_i \lambda_i^2$ (effective dimensionality) and log-volume $\log \mathrm{Vol}(\Sigma) = \sum_{i} \log \max(\lambda_i,\varepsilon)$ (neighborhood size). No truncation rank is in force for either measure: participation ratio uses the full spectrum, and the spectrum entering log-volume comes from an economy SVD of the $16\times D$ centred matrix, so it has $16$ entries, all of which are used. Both arms are estimated at identical $n$ and ambient width, so the structural offset of a $15$-dimensional sample span is common to the arms and largely differences out of the refinement-minus-control contrast (Appendix~\ref{app:n16}). All summaries use chain-level aggregation with $95\%$ confidence intervals. Recurrence across all four models is a core control on interpretation.

\section{Experiments and Results}
\subsection{A depth-local semantic structure emerges after deflation}
\label{sec:order_results}

Under cosine similarity, deflation reveals R; under Mahalanobis distance, B holds already in raw space; deflated Euclidean distance follows Mahalanobis under mean pooling and stays on the cosine side under last-token readout (Table~\ref{tab:orderings}). Under cosine, the relatedness groups overlap in raw space: strict R holds at no raw layer except two in Gemma. Deflation separates them. At the deflation level used in the figures, strict R holds at $21\%$ (Pythia), $59\%$ (Llama), $76\%$ (Qwen) and $19\%$ (Gemma) of layers under mean pooling. Under last-token readout it holds at $37$--$55\%$ of layers in Llama, Qwen and Gemma; Pythia does not reach it. In the variance-relative frame, B holds at $84$--$100\%$ of raw layers under mean pooling (Mahalanobis, $r=128$) and at $93$--$100\%$ after deflation. Under last-token readout it is weaker at the paper's retained rank: at $r=256$, B holds at $42$--$70\%$ of raw layers in Llama, Qwen and Gemma ($58$--$73\%$ after deflation), against $97\%$ in Pythia. Only at $r=512$ does it become strong in all four models, at $92$--$98\%$ of raw layers and $97$--$98\%$ after deflation. Each frame's ordering fails in the other: cosine almost never satisfies B, and Mahalanobis almost never satisfies R (Table~\ref{tab:orderings}, Appendix~\ref{app:criterion}).

The emergence of strict R is not a one-model effect. With cosine similarity under mean pooling, no layer satisfies it in the raw space except two in Gemma, and none at $k=1$--$3$; qualifying layers first appear at $k=4$ in all four models. In Llama and Qwen they then become more numerous with further deflation, reaching $24/29$ and $28/37$ layers by $k=10$. The other two models are less regular: Pythia has no qualifying layer at $k=5$--$6$ and qualifies consistently only from $k=7$ ($15/33$ by $k=10$), and Gemma's count falls from $12$ at $k=4$ to $6$ at $k=6$ before rising to $17$ at $k=10$ (Appendix~\ref{app:criterion}, Table~\ref{tab:criterion}). The deflation levels used in the figures---$k=5$ for Llama, Qwen and Gemma and $k=8$ for Pythia under mean pooling, $k=4$ for Llama, Qwen and Gemma and $k=10$ for Pythia under last-token readout---were not selected to maximize this count; Table~\ref{tab:criterion} reports the full sweep.

\begin{figure}[t]
\centering
\includegraphics[width=\textwidth]{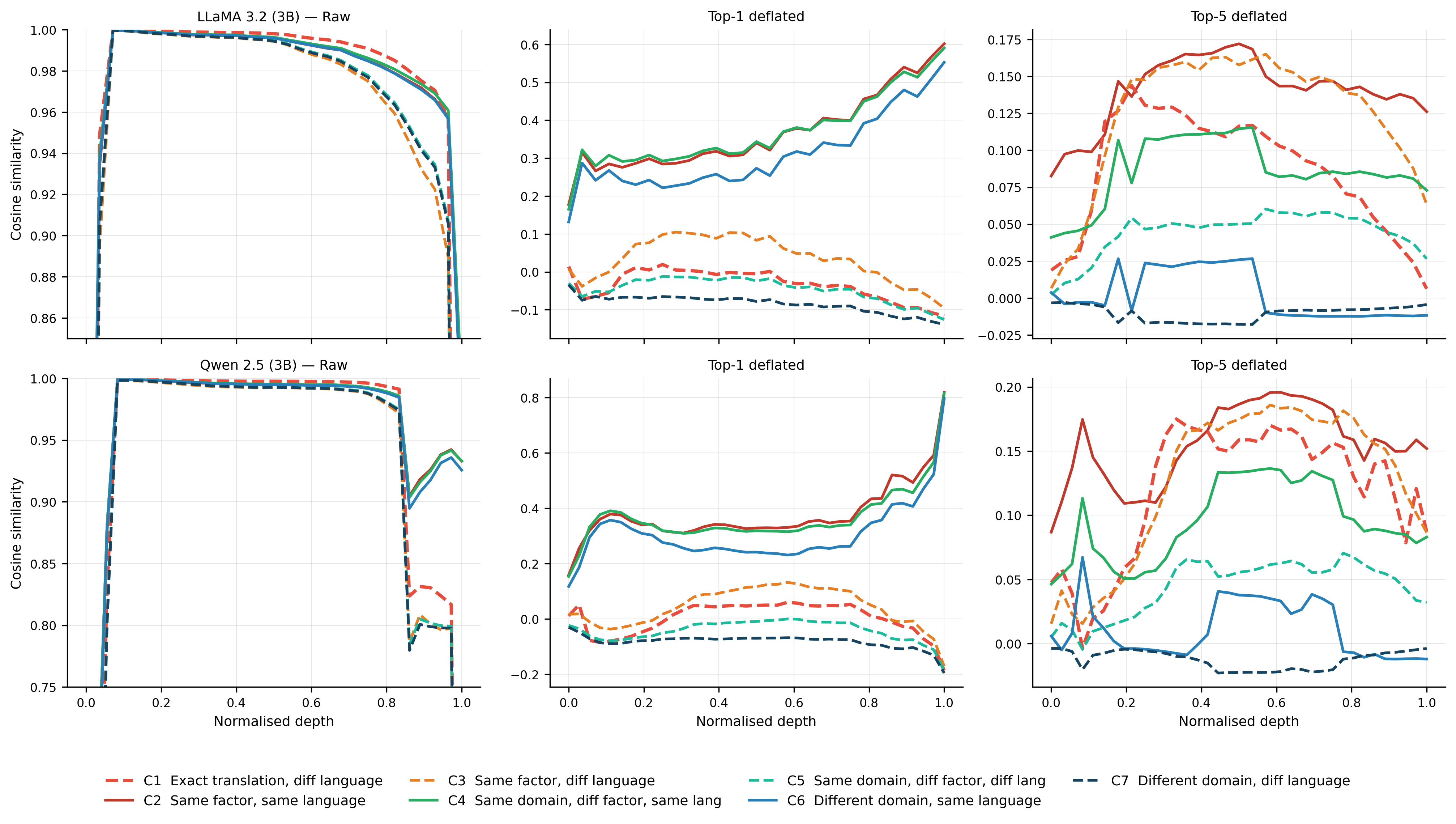}
\caption{Cosine geometry for Llama 3.2 3B (top) and Qwen 2.5 3B (bottom)
under raw, top-1, and top-5 deflation. Removing dominant directions
reveals a clearer semantic hierarchy, consistent with the
dominant-subspace masking documented in prior work
\citep{Ethayarajh2019,MuViswanath2018,CaiEtAl2021}.}
\label{fig:image1}
\end{figure}

Figure~\ref{fig:image1} shows the progression under cosine similarity.
Raw geometry is organized by language, consistent with the known dominance
of language identity in multilingual representations
\citep{Conneau2020,ChiEtAl2020}. By top-5 deflation, same-factor pairs
separate from same-domain pairs, which separate from cross-domain contrasts.
The recovered order has behavioral consequences: at normalized depth $0.5$, top-$k$
deflation improves cross-lingual Recall@1 from $0.175$ to $0.401$
(Pythia), $0.225$ to $0.608$ (Llama), $0.346$ to $0.726$ (Qwen), and
$0.220$ to $0.330$ (Gemma). Depth $0.5$ was fixed in advance and applied uniformly, so these values are conservative: it is not each model's best layer (Gemma reaches $0.700$ at depth $0.35$). The effect is not carried by the European languages: averaged over models, deflated Recall@1 is $0.62$ (English), $0.59$ (Spanish), $0.51$ (Chinese), $0.50$ (German) and $0.37$ (Arabic), and deflation improves retrieval for every language at every depth tested (Appendix~\ref{app:perlang}).

\begin{figure}[t]
\centering
\includegraphics[width=\textwidth]{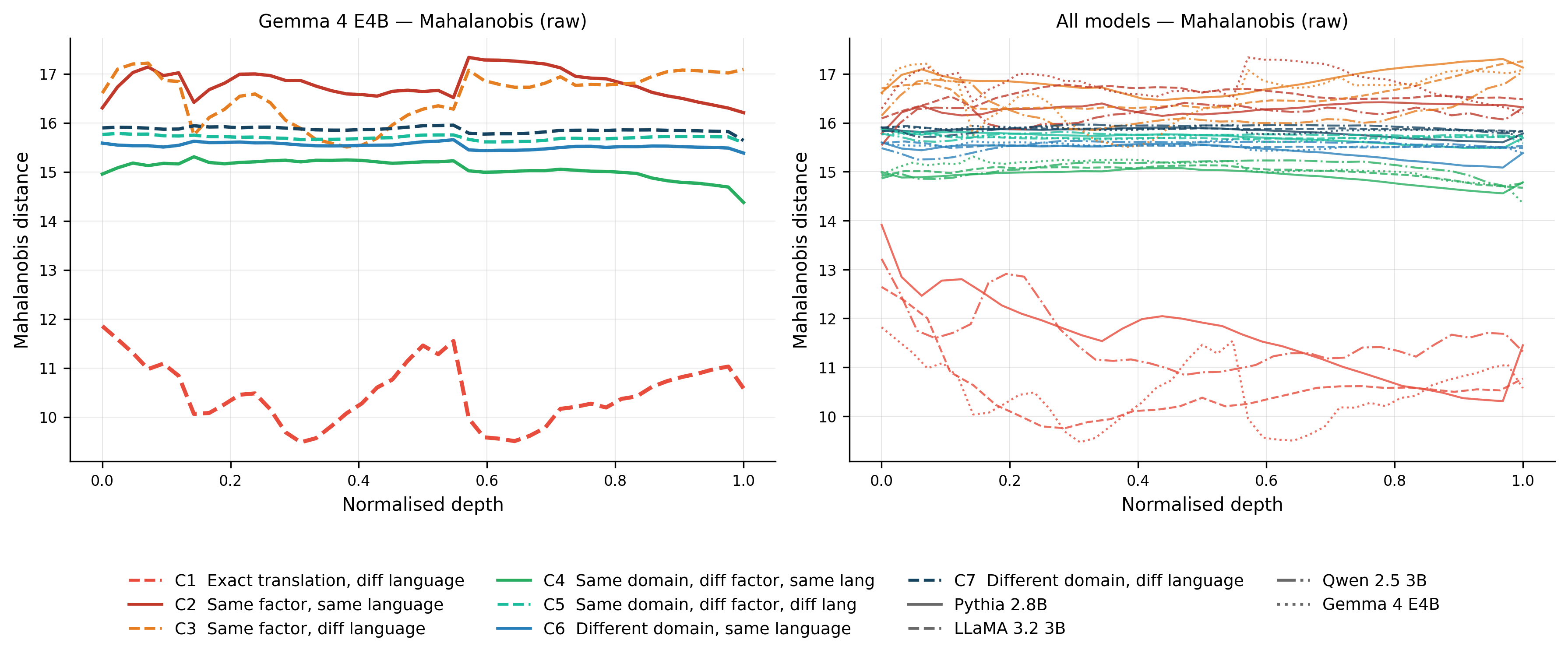}
\caption{Mahalanobis distance in the semantic-neighborhood probe. Left:
Gemma 4 E4B raw profile. Right: all-model profiles. Exact meaning matches
(C1) form the nearest category; same-factor independent realizations
(C2, C3) are the farthest, exposing the full realization spread of each
factor neighborhood relative to the ambient variance of the layer.}
\label{fig:image2}
\end{figure}

Mahalanobis distance reveals a richer ordering than cosine or Euclidean
distance alone (Figure~\ref{fig:image2}). Three features of this ordering
are consistent across all four models. First, exact meaning matches (C1)
are always the nearest category---what we call the invariant core of a semantic factor.
Second, inter-concept categories (C4--C7) occupy a compressed middle
band: same-domain and cross-domain pairs are closer together than one
might expect, because the covariance-weighted metric normalizes away the
large-scale ambient structure that separates domains in Euclidean space.
Third, same-factor independently realized pairs
(C2, C3) are the \emph{farthest} category---farther even than pairs from
different semantic domains. This is not a contradiction of the
invariant-core finding but its complement: when realization is controlled
(C1), the factor collapses to a tight region; when realization varies
freely (C2, C3), the full directional spread of the factor neighborhood
is exposed. Because Mahalanobis distance measures displacement relative to the layer's ambient variance, the finding is precisely that a factor's local spread runs along directions the ambient distribution treats as minor. In this sense a factor is not a point but an extended, anisotropic
region with a small invariant center and large realization-specific spread.
This ordering is already visible in raw Mahalanobis distance and largely
unchanged across deflation levels. It is not an artifact of the retained rank: B holds from $r=128$ upward under mean pooling in all four models (Appendix~\ref{app:rank}). When $\Sigma$ is instead estimated on disjoint factors or domains, C1 remains nearest in all $92$ configurations tested and B holds in $41$, with failures averaging a $5\%$ overlap between the middle band and C2/C3 (Appendix~\ref{app:indepcov}): the invariant core does not depend on how $\Sigma$ is estimated, while the strict separation of the outer bands is more sensitive.
The decomposition into same-facet and cross-facet same-factor
pairs is provided in Appendix~\ref{app:facets}.

\begin{figure}[t]
\centering
\includegraphics[width=\textwidth]{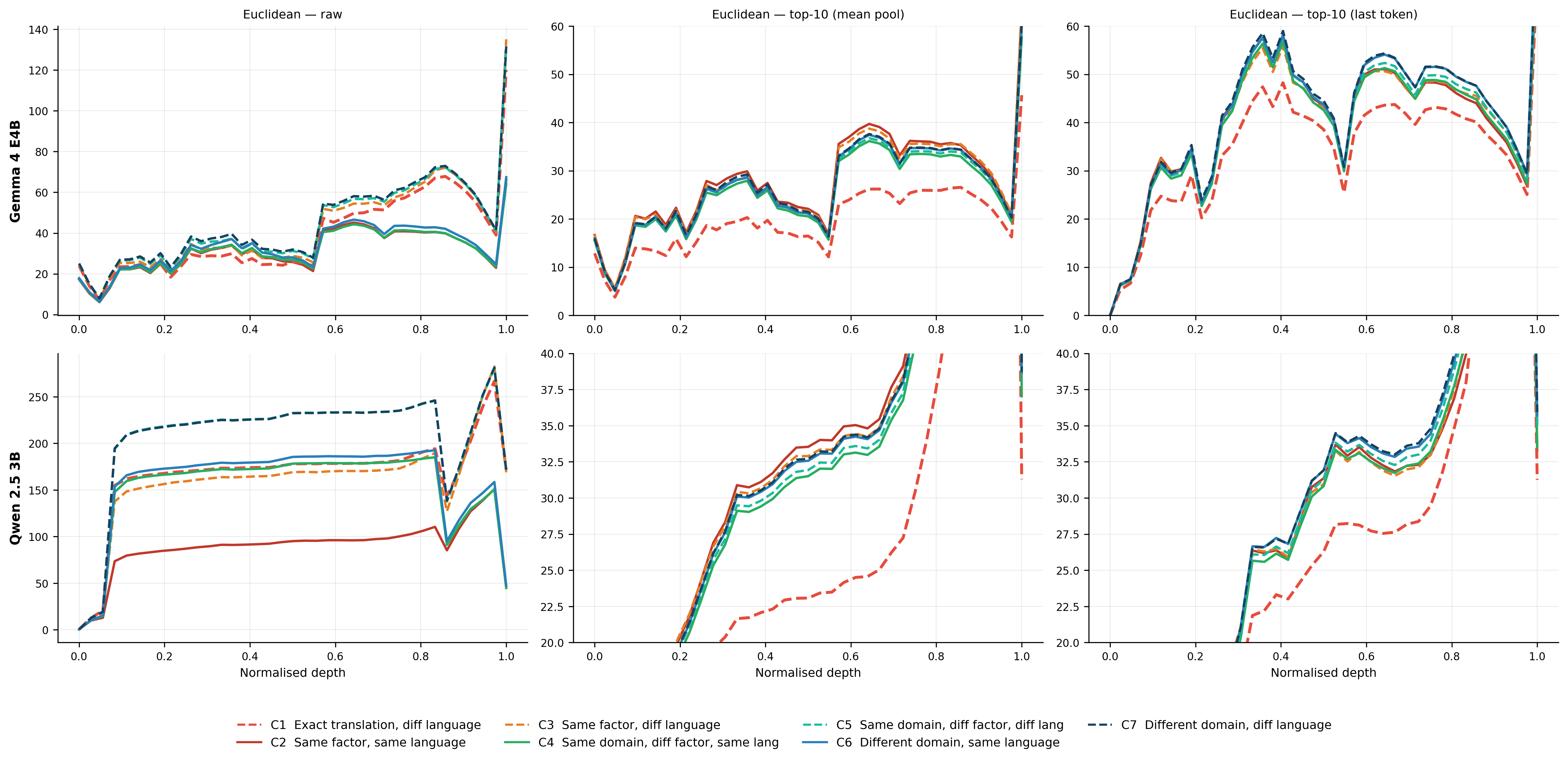}
\caption{Euclidean distance under raw and top-10-deflated conditions for
Gemma 4 E4B and Qwen 2.5 3B. In mean pooling, deflation pushes Euclidean
ordering toward the Mahalanobis ordering; in last-token pooling, it
does not move toward the Mahalanobis ordering.}
\label{fig:image3}
\end{figure}

Under deflation, Euclidean distance---which applies no inverse-variance weighting, although deflation itself is a covariance-derived spectral filter---moves progressively toward B under mean pooling as more components are removed, holding at $81$--$100\%$ of layers by top-$10$ deflation. This shows that the ordering is not specific to inverse-variance weighting; it does not establish independence from covariance-based preprocessing more broadly. Under last-token readout, deflated Euclidean distance does not make this move: like cosine, it rarely satisfies B in Llama, Qwen and Gemma, although it separates the relatedness groups less cleanly than cosine does (strict R at $5$--$21\%$ of layers, against $37$--$55\%$ for cosine). Pythia is the exception, reaching B. Both observations (Figure~\ref{fig:image3}), and a tentative account of the readout split, are reported in Appendix~\ref{app:euclid}.

\FloatBarrier
\subsection{Semantic refinement is directed displacement into tighter neighborhoods}
\label{sec:refinement_results}

The central result: the greater the semantic refinement, the farther the sentence moves from the base concept, and the more it contracts into a tighter local neighborhood. This exceeds matched control motion, survives across readouts and metrics, holds against controls built on the same base noun, and is strongest for genuine refinement rather than adversarial manipulations.

\begin{figure}[t]
\centering
\includegraphics[width=\textwidth]{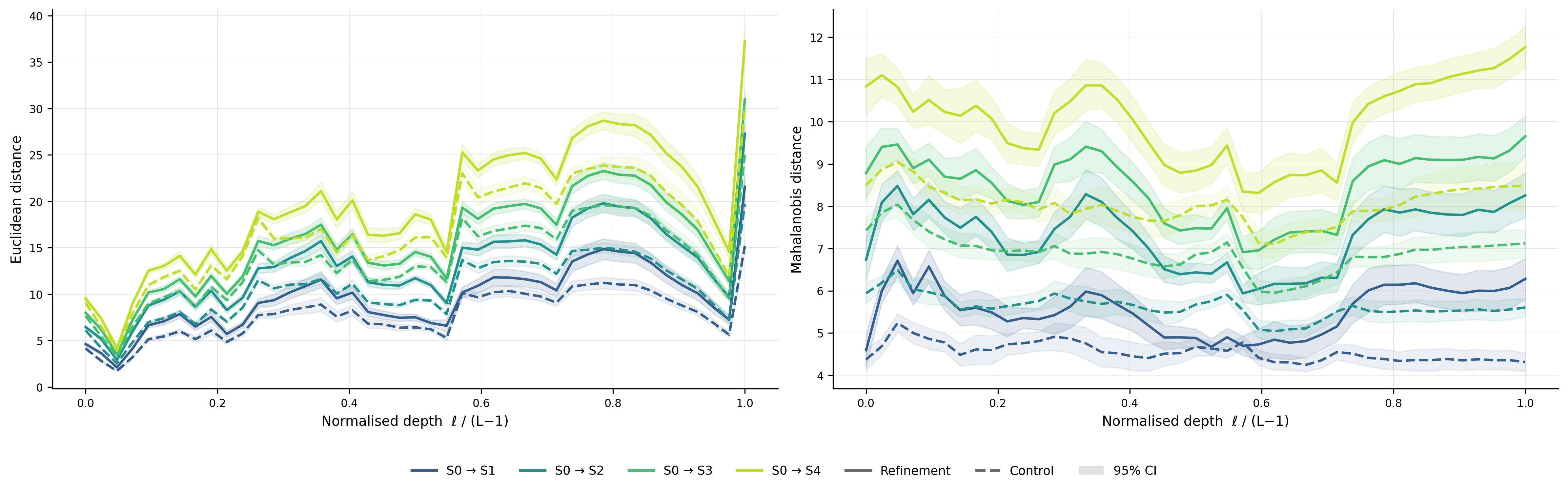}
\caption{Refinement versus control in Gemma 4 E4B under mean pooling. Distances from base concept $S_0$ to later levels (left: Euclidean; right: Mahalanobis). Refinement trajectories lie above control across most depths and target levels.}
\label{fig:image4}
\end{figure}

\textbf{Displacement.} For $S_0 \rightarrow S_4$, mean-pooled Euclidean refinement exceeds control on every layer in all four models, with strict CI separation on $21\%$ of layers in Llama, $24\%$ in Qwen, $33\%$ in Pythia, and $93\%$ in Gemma. Mean-pooled Mahalanobis refinement also exceeds control on every layer in all four models, and the strict separation rises to $98$--$100\%$. In last-token pooling, Euclidean refinement exceeds control on $94$--$98\%$ of layers with strict separation on $33$--$93\%$, while Mahalanobis refinement exceeds control on $97$--$98\%$ of layers with strict separation on $76$--$97\%$ (Figure~\ref{fig:image4}).

This effect is strong enough that partial refinement can outrun full control. In Llama last-token Mahalanobis distance, $S_0 \rightarrow S_2$ refinement is larger than $S_0 \rightarrow S_4$ control throughout the final fifth of the network. The $S_0 \rightarrow S_3$ refinement trajectory is larger still, with a late-layer mean of $9.00$ and a final-layer value of $9.10$. Semantically meaningful narrowing therefore produces more geometric displacement than simply accumulating compatible control content.

\begin{figure}[t]
\centering
\includegraphics[width=\textwidth]{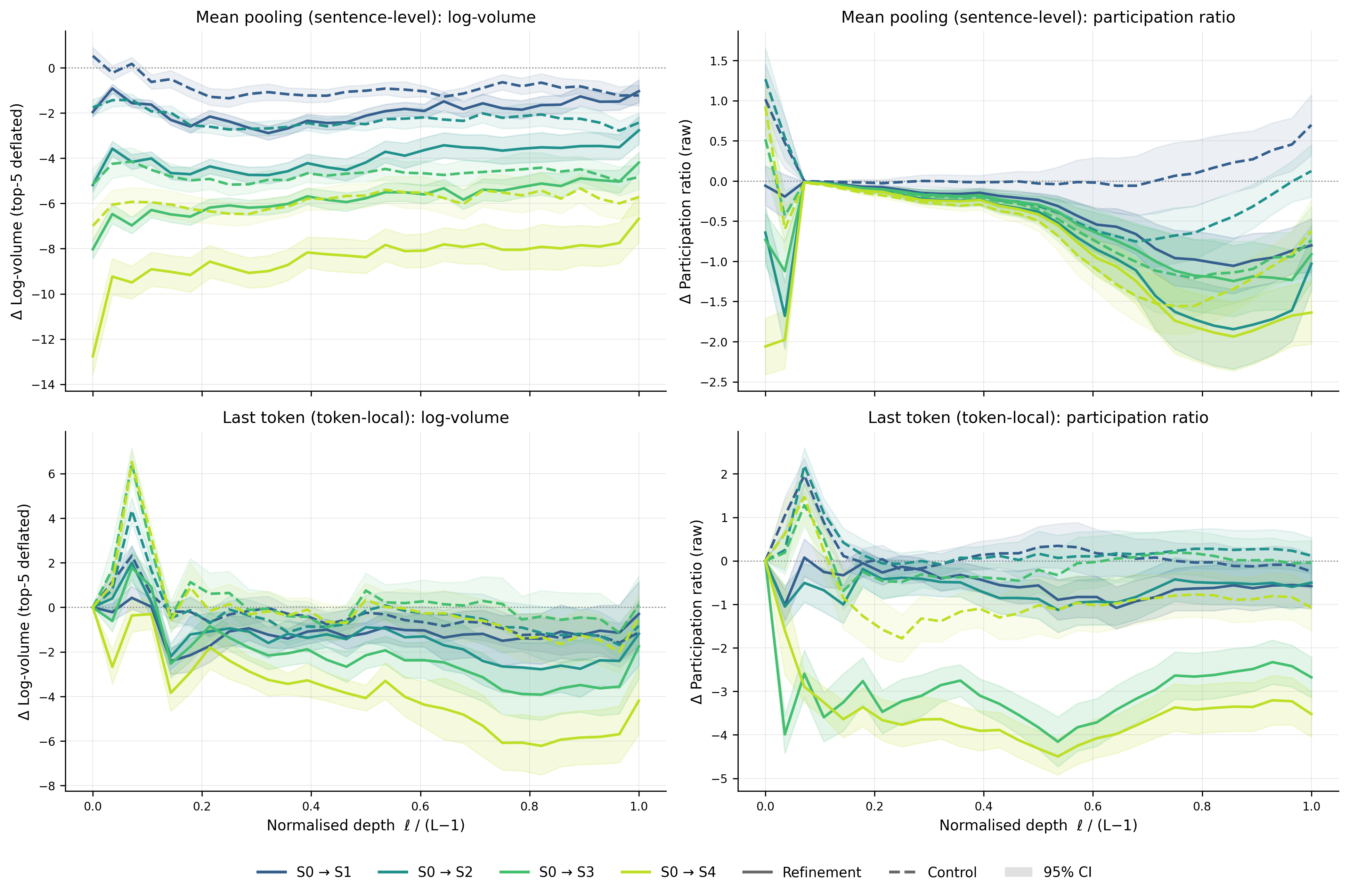}
\caption{Refinement geometry in Llama 3.2 3B, gaps from $S_0$. Rows: readout (top: mean pooling, sentence-level; bottom: last token, token-local). Columns: spread measure (left: top-5-deflated log-volume; right: raw participation ratio). Refinement (solid) contracts local volume more strongly than control (dashed) under both readouts; participation ratio separates the arms most clearly at token-local readout.}
\label{fig:image5}
\end{figure}

\textbf{Contraction.} The $S_0 \rightarrow S_4$ log-volume gap favors stronger contraction under refinement on $97$--$100\%$ of layers across models in mean pooling, with strict CI separation on $62$--$94\%$ (Figure~\ref{fig:image5}, top left). Under top-5 deflation, the same contrast remains positive on $86$--$100\%$ of layers, with strict separation on $38$--$97\%$ depending on model and span. The gap is significant under a permutation test that reshuffles arm labels across chains in all four models ($p \le 0.0024$; Appendix~\ref{app:n16}). This is the clearest neighborhood-level signature of refinement in the paper. Log-volume also separates the arms under last-token readout ($78$--$97\%$ of layers raw, $95$--$97\%$ after deflation; Table~\ref{tab:lv_summary} and Figure~\ref{fig:image5}, bottom left). In last-token readout, refinement additionally ends with \emph{lower} participation ratio than control on $97\%$ (Llama), $89\%$ (Qwen), and $70\%$ (Gemma) of layers (Figure~\ref{fig:image5}, bottom right), whereas under mean pooling the participation-ratio gap is smaller and less consistent (top right). At $n=16$ the participation-ratio estimator is strongly compressed, so we read it as an ordinal indicator rather than an estimate of effective dimensionality: log-volume carries the sentence-level contraction signature, participation ratio the token-local one. Refinement therefore contracts the sentence-level neighborhood while producing a sharper, more directionally concentrated token-local endpoint.

\textbf{Same-base comparison.} Refinement and control chains in the main analysis start from different base nouns, so the contrast could in principle reflect base-category geometry, lexical frequency or syntax. The seven base nouns that appear in both arms allow a matched comparison under an identical premodifier-stacking frame. Matching leaves contraction essentially unchanged (log-volume gap unpaired vs.\ matched: $2.28$/$2.36$ Pythia, $1.81$/$1.75$ Llama, $1.84$/$1.71$ Qwen, $2.60$/$2.07$ Gemma) and enlarges displacement (Euclidean gap $3.56$/$6.85$, $0.74$/$2.02$, $4.83$/$14.12$, $2.72$/$3.24$); within base, refinement moves farther than its matched control for a majority of the seven bases at every layer in all four models (Appendix~\ref{app:paired}). Because the seven bases share one register, we claim only that matching does not attenuate the contrast.

\textbf{Robustness to probe artifacts.} Removing every sentence with an explicit narrowing cue (e.g., ``More narrowly'', $6.3\%$ of the probe, all in one template family) or ending in a dangling function word ($2.1\%$), and comparing at matched sample size, preserves all three qualitative Figure~\ref{fig:image5} claims (refinement below control at $97\%$ of layers, against $98\%$ unfiltered; monotone ordering by level and $S_4$ contraction at $100\%$), with displacement sign agreement of $100\%$ on every layer in all four models (Appendix~\ref{app:qc}).

\textbf{Adversarial controls.} Near-synonym stacking and scope hedging serve as a stress test, not as primary evidence, that the refinement--control contrast does not simply reflect control's construction: if the pattern were definitional, any arm built under a different rule would diverge from refinement in the same direction. It does not do so uniformly. In participation-ratio terms, refinement reaches negative peak gaps in all four models ($-0.12$, $-0.04$, $-0.08$, $-0.15$), while the adversarial controls are mixed and usually near zero or positive; in log-volume terms, scope-hedge is consistently the shallowest arm across all models, and near-synonym is less negative than refinement in two of four. Each adversarial arm contains only four chains, and the near-synonym results are mixed. Full depth profiles with 95\% CI bands appear in Appendix~\ref{sec:adv_appendix}.

\begin{figure}[t]
\centering
\includegraphics[width=\textwidth]{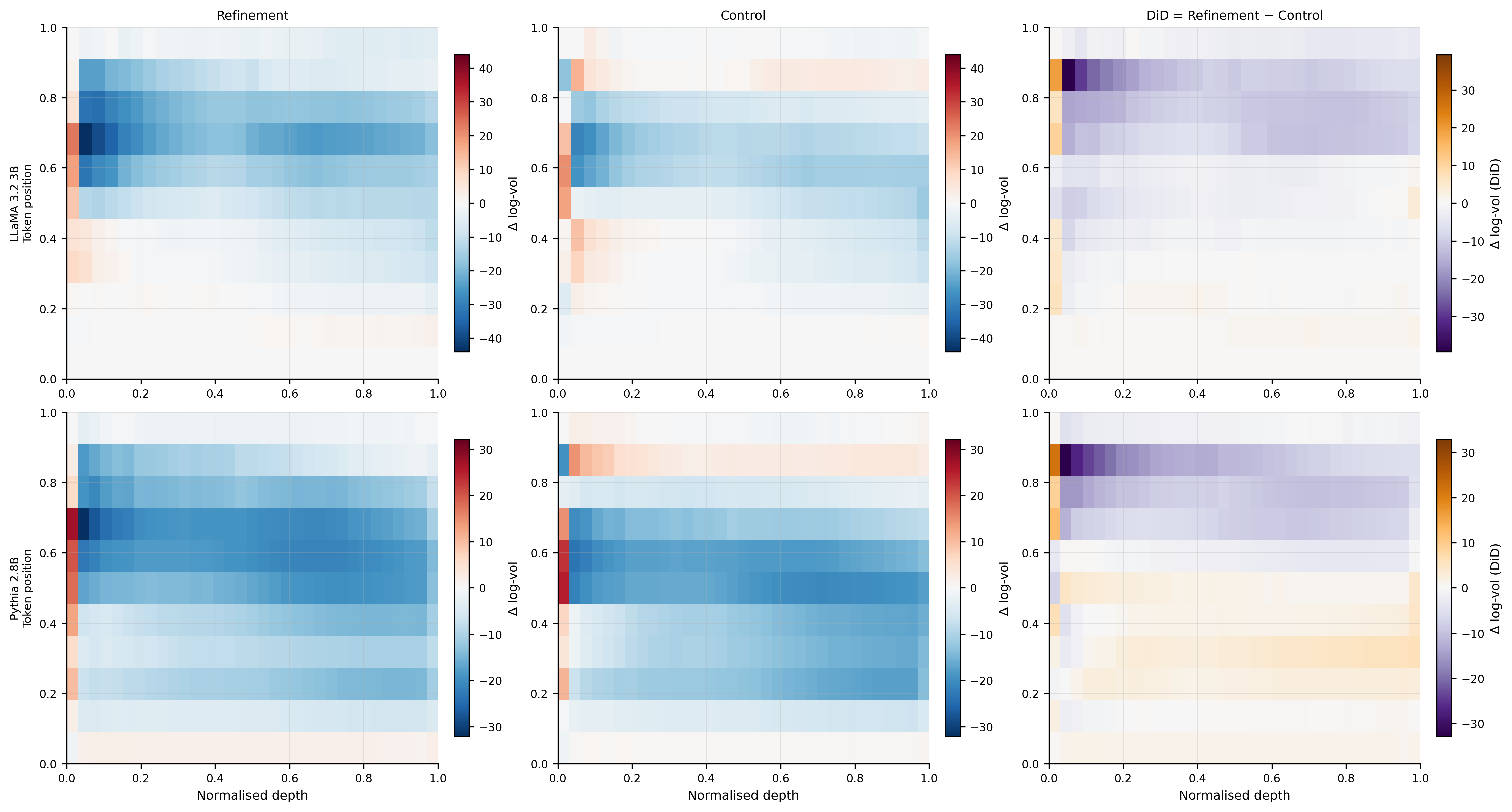}
\caption{Per-position increments in log-volume (Llama 3.2 3B top, Pythia 2.8B bottom); columns show refinement, control, and their difference. Refinement-specific contraction strengthens through late-prefix positions; at the terminal period the increment is near zero in both arms.}
\label{fig:image6}
\end{figure}

\textbf{Per-position analysis.} Tracking the per-position \emph{increment} in log-volume shows that neighborhood contraction grows across late-prefix positions and is stronger for refinement than for control, with log-volume gaps over positions $0.7$--$0.9$ averaging $-9.0$ (Pythia), $-10.6$ (Llama), $-12.9$ (Qwen). At the terminal period the increment is near zero in both arms, and the refinement-minus-control difference collapses ($-1.29$, $-2.60$, $-1.26$). The period's hidden state attends to the whole prefix and is not informationally empty; a near-zero increment at punctuation admits several explanations, so we treat it as an observation consistent with the prefix-level account rather than as evidence for it. The per-position maps are shown in Figure~\ref{fig:image6}.

\section{Discussion and Conclusion}

\paragraph{A unified geometric picture.}
The two probes tell one story. In the semantic-neighborhood probe, cosine geometry orders pairs by relatedness; variance-relative geometry shows a tight core and wide realization spread. They are two views of the same neighborhoods: exact meaning matches occupy the most invariant region, and same-factor descriptions retain structured local variation. The refinement probe shows the dynamic complement: narrowing meaning produces directed displacement into tighter neighborhoods. We interpret this as LLM semantics being organized as local anisotropic neighborhoods with an invariant core, realization-specific directions, and inward refinement trajectories.

\paragraph{Relation to the linear representation hypothesis.}
The linear representation hypothesis \citep{ParkEtAl2024,ElhageEtAl2022} identifies directions as carriers of meaning. The present findings are complementary: linear directions provide the backbone, but neighborhoods around those directions have additional structure---an invariant core, anisotropic spread, and contraction under refinement---that a single direction does not capture. The participation-ratio results suggest that this structure is not isotropic even locally, which may bear on how steering vectors \citep{TurnerEtAl2023,ZouEtAl2023} should be calibrated.

\paragraph{Possible implications for representation-based methods.}
Methods that intervene on hidden states---steering vectors \citep{TurnerEtAl2023,ZouEtAl2023}, inference-time intervention \citep{LiEtAl2023ITI}, and latent-space classifiers \citep{BurnsEtAl2023,MarksTegmark2024}---implicitly assume directional geometry. The present results suggest that local spread and contraction provide additional geometric signals that could serve as complementary diagnostics. For example, if genuine semantic narrowing reliably produces volume contraction and directional concentration, then monitoring these signatures could help distinguish semantic specialization from qualitatively different representational shifts---a question relevant to alignment monitoring. Conversely, a steering vector that points in the right direction but ignores local anisotropy may land in a different part of the neighborhood than intended. These applications are not tested here; the contribution is the characterization on which such tests could be built.

\paragraph{Cross-architecture replication.}
The replication across four model families---Pythia (GPT-NeoX, the Pile \citep{BidermanEtAl2023}), Llama (grouped-query attention, proprietary corpus \citep{GrattafioriEtAl2024}), Qwen (distinct BPE, different language mixture \citep{YangEtAl2024Qwen}), and Gemma (different depth-width tradeoff \citep{GemmaTeam2026})---that share no tokenizer or architectural template suggests that the findings reflect a property of autoregressive transformer semantics in this size range rather than an idiosyncrasy of any single design.

\paragraph{Limitations.}
First, all models are in the 3--5B range; whether the geometric signatures persist, sharpen, or change qualitatively at 70B+ is an open empirical question. Second, the analyses characterize geometry but do not manipulate it; the bridge from geometric characterization to causal intervention---using volume contraction or anisotropy as control signals---remains untested. Third, the local spread measures are estimated from $n=16$ realizations per chain-level in a $2{,}048$--$3{,}072$-dimensional space, so the claims rest on refinement-minus-control contrasts at matched $n$ rather than on absolute values, and participation ratio is read as an ordinal indicator. Fourth, the control arm also narrows its base noun in a weak sense, so the contrast is a conservative lower bound; the same-base comparison rests on seven nouns from one register, and the adversarial arms contain four chains each.

\paragraph{Conclusion.}
Taken together, the evidence supports one geometric interpretation:
hidden states contain a semantic order over a band of layers that becomes visible
after dominant directions are removed; semantic factors combine an
invariant core with realization-specific local variation; and refinement
is directed motion into tighter semantic neighborhoods. At the
token-local readout, the stronger the semantic
narrowing, the more the final token concentrates into fewer directions.
These findings hold across four
architecturally diverse families and offer a complementary geometric
description---beside the linear representation hypothesis---for
understanding meaning in language models.

\bibliographystyle{plainnat}
\bibliography{references}

\appendix

\section{Same-Facet and Cross-Facet Same-Factor Pairs}
\label{app:facets}

The broad same-factor cross-lingual category in the semantic-neighborhood probe admits a more precise split than the main text needs. Exact translations remain their own category: they preserve both meaning and parallel sentence slot across languages. The independent same-factor realizations can then be separated into \emph{same-facet} pairs, which preserve the same semantic facet across languages, and \emph{cross-facet} pairs, which preserve the factor while changing the facet used to express it. This distinction matters because cross-facet does not mean error, noise, or mistranslation. It means that the same factor is being realized through a different semantic route.

\begin{table}[H]
\centering
\small
\caption{Examples distinguishing exact translation from same-facet and cross-facet same-factor pairs for the factor \emph{hot}.}
\label{tab:facet_examples}
\resizebox{\textwidth}{!}{%
\begin{tabular}{p{3.4cm}p{5.0cm}p{5.0cm}}
\toprule
Pair type & English & German \\
\midrule
Exact translation & The object is hot. & Das Objekt ist hei{\ss}. \\
Same-facet same-factor & It has a very high temperature. & Es hat eine sehr hohe Temperatur. \\
Cross-facet same-factor & You would burn yourself on it. & Man w\"urde sich daran verbrennen. \\
\bottomrule
\end{tabular}
}
\end{table}

Table~\ref{tab:facet_examples} illustrates the logic. The exact-translation category isolates sentence-level identity across languages. The same-facet category stays within the independent-realization pool but preserves the same semantic facet. The cross-facet category keeps the factor fixed while moving from one facet of the neighborhood to another, for example from a direct temperature statement to a consequence-based realization.

The empirical split is summarized in Table~\ref{tab:facet_dm}. Across models and readouts, same-facet pairs are already farther than exact matches, and cross-facet pairs are farther still. The cross-facet distance is $1.01$--$1.37\times$ the same-facet distance, while the same-facet pairs remain $1.04$--$1.37\times$ farther than exact translations. The nearest region in Mahalanobis geometry is therefore not the full factor neighborhood but an invariant semantic center surrounded by structured realization-specific spread.

\begin{table}[H]
\centering
\small
\caption{Mahalanobis means for exact matches, same-facet same-factor pairs, and cross-facet same-factor pairs inside the model-specific semantic band. Cross-facet pairs are consistently farther than same-facet pairs.}
\label{tab:facet_dm}
\resizebox{\textwidth}{!}{%
\begin{tabular}{llcccc}
\toprule
Model & Readout & Exact match $d_M$ & Same-facet $d_M$ & Cross-facet $d_M$ & Cross-facet / same-facet \\
\midrule
Pythia 2.8B & Mean & 11.54 & 15.36 & 17.18 & 1.12 \\
Pythia 2.8B & Last-token & 12.42 & 16.90 & 17.14 & 1.01 \\
Llama 3.2 3B & Mean & 9.57 & 13.10 & 17.37 & 1.33 \\
Llama 3.2 3B & Last-token & 12.60 & 13.14 & 16.15 & 1.23 \\
Qwen 2.5 3B & Mean & 10.82 & 13.34 & 16.88 & 1.27 \\
Qwen 2.5 3B & Last-token & 12.53 & 13.13 & 16.13 & 1.23 \\
Gemma 4 E4B & Mean & 9.99 & 12.66 & 17.33 & 1.37 \\
Gemma 4 E4B & Last-token & 13.01 & 14.37 & 15.90 & 1.11 \\
\bottomrule
\end{tabular}
}
\end{table}

\section{Dataset Construction Examples}

The tables below provide concrete examples of the pair categories and refinement chains used throughout the paper.

\begin{table}[H]
\centering
\small
\caption{Examples of the seven semantic-neighborhood pair categories.}
\label{tab:pair_examples}
\resizebox{\textwidth}{!}{%
\begin{tabular}{lp{7cm}}
\toprule
Category & Example (\emph{The object is hot.} $\rightarrow$) \\
\midrule
C1 Exact meaning, cross-language & Das Objekt ist hei{\ss}. \\
C2 Same factor, same language & It has a very high temperature. \\
C3 Same factor, cross-language & \\
\quad Same facet & Es hat eine sehr hohe Temperatur. \\
\quad Cross-facet & Man w\"urde sich daran verbrennen. \\
C4 Same domain, different factor, same language & The object is cold. \\
C5 Same domain, different factor, cross-language & Das Objekt ist kalt. \\
C6 Different domain, same language & Its surface has noticeable irregularities. \\
C7 Different domain, cross-language & Die Oberfl\"ache f\"uhlt sich rau an. \\
\bottomrule
\end{tabular}}
\end{table}

The semantic-neighborhood categories make it possible to distinguish exact identity, same-factor variation, nearby concepts, and clear domain shifts while holding sentence style tightly controlled. The refinement chains make the second probe interpretable for the same reason: the refinement arm narrows meaning, the main control adds compatible description without committing to a marked subtype, and the adversarial controls change wording or scope without implementing the same narrowing operation. The first two rows of Table~\ref{tab:chain_examples} are a same-base pair.

\begin{table}[H]
\centering
\small
\caption{Example chains from the semantic refinement probe. The first two rows share the base noun \emph{report}.}
\label{tab:chain_examples}
\resizebox{\textwidth}{!}{%
\begin{tabular}{llllll}
\toprule
Arm & $S_0$ & $S_1$ & $S_2$ & $S_3$ & $S_4$ \\
\midrule
Refinement & report & audit report & environmental audit report & annual environmental audit report & certified annual environmental audit report \\
Main control & report & written report & written dated report & written dated formal report & written dated formal official report \\
\midrule
Refinement & chair & office chair & ergonomic office chair & adjustable ergonomic office chair & adjustable ergonomic mesh office chair \\
Main control & document & written document & written dated document & written dated signed document & written dated signed official document \\
Near-synonym control & object & large object & large big object & large big huge object & large big huge enormous object \\
Scope-hedge control & object & fragile object & partly fragile object & somewhat partly fragile object & broadly somewhat partly fragile object \\
\bottomrule
\end{tabular}}
\end{table}

\section{Human Validation of the Probes}
\label{app:human}

Three blind studies were run after the probes were frozen, with annotators independent of the project (volunteer colleagues and university staff, none an author), none of whom was shown any model output, figure or hypothesis.

\paragraph{Study 1: paired comparison (refinement probe).}
Three native English speakers each made $60$ forced choices. Each item shows two pairs of phrases; in each pair a short starting term grows into a longer one, and the annotator decides which longer term is more tightly pinned down, by picturing the set of things each starting term could refer to and how much of that set survives once the longer term is used. Sides were counterbalanced $30/30$, comparisons were base- or register-matched, and nothing indicated provenance. Annotators matched the intended ordering on $179$ of $180$ comparisons ($99.4\%$, $95\%$ CI $96.9$--$99.9$), Fleiss' $\kappa = 0.978$, with no decay across chain depth.

\paragraph{Study 2: graded step judgment (refinement probe).}
As a harder test, three further annotators each judged $160$ isolated steps (e.g., \emph{document} $\rightarrow$ \emph{written document}) on a four-point scale from ``clearly more specific'' to ``already assumed''. The scale definitions were the entire instruction, with no worked examples and no indication of expected frequencies; embedded attention checks were all passed. Across $480$ judgments, steps were assigned to the correct chain type $82.3\%$ of the time ($95\%$ CI $78.6$--$85.4$), Cohen's $\kappa = 0.646$ against our labels, odds ratio $21.6$, symmetric across arms (refinement $82.1\%$, control $82.5\%$). Since displacement and contraction are computed per chain, the chain is the relevant unit: taking each annotator's majority verdict per chain, $17$ of $20$ refinement chains were confirmed, together with all $20$ control chains. The three exceptions are \emph{blanket}, \emph{camera} and \emph{toy}.

\paragraph{Study 3: cross-lingual identification (semantic-neighborhood probe).}
Twelve submissions from native speakers across the four non-English languages, $40$ items each. Each item shows one English sentence and two target-language sentences judged separately, one a foil and one a genuine realization. Agreement with our factor assignments ranged from $88.8\%$ to $100\%$ across languages and question types, pooling to $94.5\%$ across $960$ judgments ($95\%$ CI $92.8$--$95.8$). Inter-annotator agreement was Cohen's $\kappa = 0.889$ on close parallels and $0.723$ on independent realizations; where lower, both annotators in the weakest pair still rejected foils near-perfectly ($13$ of $14$ same-domain, $6$ of $6$ cross-domain), differing only on how far an indirect expression may drift and still count as the same property. All $40$ refinement and control chains, all $160$ level transitions and all $40$ factors in every non-English language were covered.

\section{Two Orderings: Detailed Results}
\label{app:criterion}

The two-orderings results of Table~\ref{tab:orderings} are computed from one exact SVD per layer, which yields the deflated states and the Mahalanobis eigenspectrum together; the per-model values are released with the code, and the cosine counts reproduce Table~\ref{tab:criterion} exactly.

\begin{table}[H]
\centering
\caption{Two orderings across metrics and readouts: range over the four models of the percentage of layers satisfying strict R and B, in raw space $\rightarrow$ at the deflation level used in the figures ($k=5$, or $8$ for Pythia, under mean pooling; $k=4$, or $10$ for Pythia, under last-token readout). Mahalanobis uses $r=128$ (mean) and $r=256$ (last-token). Cosine almost never satisfies B and Mahalanobis almost never satisfies R. Euclidean B depends on $k$ (Table~\ref{tab:eucl_sweep}); per-model values are released with the code.}
\label{tab:orderings}
\begin{tabular}{llcc}
\toprule
Readout & Metric & Strict R (\%) & B (\%) \\
\midrule
Mean & Cosine & 0--5 $\rightarrow$ 19--76 & 0 $\rightarrow$ 0 \\
Mean & Euclidean & 0--5 $\rightarrow$ 0 & 0 $\rightarrow$ 0--97 \\
Mean & Mahalanobis & 0 $\rightarrow$ 0 & 84--100 $\rightarrow$ 93--100 \\
\midrule
Last-token & Cosine & 0--12 $\rightarrow$ 0--55 & 0 $\rightarrow$ 0--3 \\
Last-token & Euclidean & 0--12 $\rightarrow$ 0--21 & 0 $\rightarrow$ 0--94 \\
Last-token & Mahalanobis & 0 $\rightarrow$ 0--3 & 42--97 $\rightarrow$ 58--97 \\
\bottomrule
\end{tabular}
\end{table}

Table~\ref{tab:criterion} reports, for cosine similarity under mean pooling, the number of layers satisfying strict R at each deflation level. Qualifying layers first appear at $k=4$ in all four models. The count grows with $k$ in Llama and Qwen; Pythia has none at $k=5$--$6$ and qualifies consistently only from $k=7$; Gemma's count falls from $12$ to $6$ between $k=4$ and $k=6$ before rising to $17$ at $k=10$.

\begin{table}[H]
\centering
\small
\caption{Layers satisfying strict R under cosine similarity (mean pooling), by deflation level $k$. Total layer count in parentheses.}
\label{tab:criterion}
\begin{tabular}{lccccccccc}
\toprule
Model & raw & $k{=}1$--$3$ & $k{=}4$ & $k{=}5$ & $k{=}6$ & $k{=}7$ & $k{=}8$ & $k{=}9$ & $k{=}10$ \\
\midrule
Pythia 2.8B (33) & 0 & 0 & 2 & 0 & 0 & 4 & 7 & 8 & 15 \\
Llama 3.2 3B (29) & 0 & 0 & 6 & 17 & 19 & 21 & 22 & 22 & 24 \\
Qwen 2.5 3B (37) & 0 & 0 & 17 & 28 & 28 & 28 & 27 & 27 & 28 \\
Gemma 4 E4B (43) & 2 & 0 & 12 & 8 & 6 & 9 & 8 & 15 & 17 \\
\bottomrule
\end{tabular}
\end{table}

\section{Sensitivity to the Retained Mahalanobis Rank}
\label{app:rank}

The rank sweep uses an exact eigendecomposition of the layer covariance at each model's own deflation level, at normalized depth $0.5$. At each rank $r$ we record whether B holds and the Spearman $\rho$ of the full seven-category ordering against an $r=512$ reference (Table~\ref{tab:rank}). Under mean pooling the criterion holds at the paper's $r=128$ in all four models and at every larger rank tested up to $r=512$; Pythia enters the plateau already at $r=64$, and below $r=64$ the ordering has not formed and $\rho$ falls toward zero. Deviations of $\rho$ from unity at $r=128$ are adjacent swaps inside the compressed middle band that leave B intact. An artifact of low-rank whitening would be strongest at small $r$ and dissolve as $r$ grows; the observed profile is the opposite. Under last-token readout the criterion fails at $r=128$ for Llama, Qwen and Gemma; it holds from $r=256$ for Llama and Qwen, and from $r=512$ for Gemma. As $r$ approaches the number of sentences ($N=1{,}200$) the covariance tail becomes unstable and $\rho$ degrades, which is the reason for truncating.

\begin{table}[H]
\centering
\small
\caption{Rank sensitivity at normalized depth $0.5$, at each model's deflation level. $\rho$: Spearman correlation of the seven-category ordering with the $r=512$ reference. \checkmark/--: B holds / fails.}
\label{tab:rank}
\resizebox{\textwidth}{!}{%
\begin{tabular}{llcccccc}
\toprule
Model & Readout & $k$ & $\rho_{r=128}$ & $\rho_{r=256}$ & B $r{=}128$ & B $r{=}256$ & B $r{=}512$ \\
\midrule
Pythia 2.8B & Mean & 8 & 1.00 & 1.00 & \checkmark & \checkmark & \checkmark \\
Llama 3.2 3B & Mean & 5 & 0.96 & 0.96 & \checkmark & \checkmark & \checkmark \\
Qwen 2.5 3B & Mean & 5 & 0.96 & 1.00 & \checkmark & \checkmark & \checkmark \\
Gemma 4 E4B & Mean & 5 & 1.00 & 1.00 & \checkmark & \checkmark & \checkmark \\
\midrule
Pythia 2.8B & Last-token & 10 & 1.00 & 0.96 & \checkmark & \checkmark & \checkmark \\
Llama 3.2 3B & Last-token & 4 & 0.71 & 0.96 & -- & \checkmark & \checkmark \\
Qwen 2.5 3B & Last-token & 4 & 0.79 & 1.00 & -- & \checkmark & \checkmark \\
Gemma 4 E4B & Last-token & 4 & 0.46 & 0.79 & -- & -- & \checkmark \\
\bottomrule
\end{tabular}}
\end{table}

The deflation level $k$ is likewise not tuned to the criterion: Section~\ref{sec:order_method} sweeps $k=1,\dots,10$, and Table~\ref{tab:criterion} reports the qualifying-layer count at every level.

\section{Covariance Estimated Independently of the Analyzed Sentences}
\label{app:indepcov}

The concern is that a covariance estimated from the analyzed sentences could induce the ordering it is used to measure. Note first what does not depend on $\Sigma$: the cosine ordering, its behavioral validation by cosine nearest-neighbor retrieval, and the refinement displacement contrast under Euclidean distance. For the Mahalanobis ordering we ran two estimation-side checks (mean pooling, normalized depth $0.5$). Besides B itself, we report a weaker \emph{band-means} check, which compares only the means of the three bands: $\bar d_{C1} < \operatorname{mean}(\bar d_{C4},\dots,\bar d_{C7}) < \operatorname{mean}(\bar d_{C2},\bar d_{C3})$. B additionally requires that no individual middle-band category overlap C1 or C2/C3.

\emph{Disjoint sentences.} Splitting the $40$ factors into halves of $20$, each covering all $10$ domains and $5$ languages, estimating $\Sigma$ on one half and measuring only pairs from the other (so that no sentence contributes to both), C1 remains nearest and the band means remain ordered in all four models, and B holds in three of the four; the seven-category ranking agrees with the full-data ranking at mean $\rho=0.97$ (lowest $0.89$). Estimating on five of the ten domains and measuring on the other five again keeps C1 nearest and the band means ordered in all four models, but B holds in none.

\emph{A different corpus.} Replacing only the covariance with one estimated from the refinement probe's $3{,}840$ sentences---differently constructed and three times larger---keeps C1 nearest in all eight model--readout cases.

Across the $92$ mean-pooled configurations run, C1 was nearest in $92/92$ and B held in $41/92$; where B fails, the overlap between the middle band and C2/C3 averages $5\%$ of the C1-to-C2/C3 range, and the band means remain ordered in all $92$. The margin that moves is narrow under the paper's own covariance too ($6$--$17\%$ of the same range). The invariant core is therefore robust to how $\Sigma$ is estimated, while the clean separation of the middle band from C2/C3 is the sensitive part of B.

\section{Per-Language Cross-Lingual Retrieval}
\label{app:perlang}

Table~\ref{tab:perlang} breaks Recall@1 down by language (averaged over the language as source and as target) at the paper's operating point (normalized depth $0.5$; chance $0.025$). Chinese performs on a par with German, so the deflation gain is not a European-script effect; Arabic is the weakest language in every model. Within each model all five languages peak at the same depth, so the peak is a property of the model rather than a per-language choice; at its best depth Arabic reaches $0.647$ (Llama), $0.666$ (Qwen) and $0.819$ (Gemma), $26$--$33\times$ chance. Pythia is weak throughout.

\begin{table}[H]
\centering
\small
\caption{Per-language cross-lingual Recall@1 at normalized depth $0.5$, raw $\rightarrow$ deflated.}
\label{tab:perlang}
\begin{tabular}{lccccc}
\toprule
Model & en & de & es & ar & zh \\
\midrule
Pythia 2.8B  & 0.29 $\rightarrow$ 0.55 & 0.18 $\rightarrow$ 0.36 & 0.18 $\rightarrow$ 0.49 & 0.06 $\rightarrow$ 0.18 & 0.16 $\rightarrow$ 0.43 \\
Llama 3.2 3B & 0.30 $\rightarrow$ 0.70 & 0.32 $\rightarrow$ 0.60 & 0.28 $\rightarrow$ 0.69 & 0.10 $\rightarrow$ 0.44 & 0.12 $\rightarrow$ 0.61 \\
Qwen 2.5 3B  & 0.41 $\rightarrow$ 0.80 & 0.39 $\rightarrow$ 0.73 & 0.38 $\rightarrow$ 0.79 & 0.24 $\rightarrow$ 0.59 & 0.31 $\rightarrow$ 0.71 \\
Gemma 4 E4B  & 0.30 $\rightarrow$ 0.43 & 0.23 $\rightarrow$ 0.32 & 0.27 $\rightarrow$ 0.38 & 0.13 $\rightarrow$ 0.26 & 0.17 $\rightarrow$ 0.27 \\
\midrule
Mean         & 0.33 $\rightarrow$ 0.62 & 0.28 $\rightarrow$ 0.50 & 0.28 $\rightarrow$ 0.59 & 0.13 $\rightarrow$ 0.37 & 0.19 $\rightarrow$ 0.51 \\
\bottomrule
\end{tabular}
\end{table}

\section{Local Spread Measures at $n=16$}
\label{app:n16}

Each local covariance is estimated from $16$ realizations for one chain at one level in one layer. The spectrum of the $16\times D$ centred matrix has $16$ entries, one of them numerically zero because of centering; log-volume sums all of them, so no $\varepsilon$-floored ambient directions enter, and a nominal rank bound of $128$ returns values identical to the natural-length call.

Two components bear on whether the refinement-versus-control gap is unbiased. The structural offset differences out: both arms are estimated at identical $n$ and identical ambient width, so both lie in a $15$-dimensional sample span with the same directions omitted. The residual is not exactly zero---including or excluding the near-zero centering eigenvalue shifts the $S_0\rightarrow S_4$ contrast by $1$--$16\%$---but the sign is preserved in all four models. Chain-level variability is the part no differencing removes; Table~\ref{tab:perm} reports a permutation test that reshuffles arm labels across the $40$ chains $10{,}000$ times at a mid-depth layer.

\begin{table}[H]
\centering
\small
\caption{Permutation $p$-values ($10{,}000$ arm-label reshuffles across chains) for the $S_0\rightarrow S_4$ refinement-minus-control gap, mean pooling, at one mid-depth layer per model.}
\label{tab:perm}
\begin{tabular}{lcccc}
\toprule
Quantity & Pythia 2.8B & Llama 3.2 3B & Qwen 2.5 3B & Gemma 4 E4B \\
\midrule
Log-volume, raw & 0.0007 & 0.0024 & 0.0005 & $<$0.0001 \\
Log-volume, top-5 deflated & $<$0.0001 & $<$0.0001 & 0.0001 & $<$0.0001 \\
Participation ratio, raw & 0.0004 & 0.164 & 0.412 & 0.005 \\
\bottomrule
\end{tabular}
\end{table}

Participation ratio is the weaker measure at this sample size: in synthetic calibration at $n=16$ the estimator is severely compressed ($10.7$ for a true value of $40.0$). It remains monotone, so we use it as an ordinal indicator and assign the sentence-level contraction claim to log-volume.

\section{Same-Base Paired Chains}
\label{app:paired}

Seven base nouns appear in both arms (\emph{certificate, form, invoice, license, notice, record, report}). Each refinement chain is compared against the control chain built on the identical noun under the same premodifier-stacking frame ($S_0\rightarrow S_4$, mean pooling, every layer). Table~\ref{tab:paired} gives the gaps in the convention of Tables~\ref{tab:disp_summary} and~\ref{tab:lv_summary}, together with the within-base vote.

\begin{table}[H]
\centering
\small
\caption{Unpaired (all $20$ vs.\ $20$ chains) versus same-base matched ($7$ vs.\ $7$) refinement-minus-control gaps, and the share of layers at which a majority of the seven bases favor refinement.}
\label{tab:paired}
\resizebox{\textwidth}{!}{%
\begin{tabular}{lcccccc}
\toprule
 & \multicolumn{2}{c}{Log-volume gap} & \multicolumn{2}{c}{Euclidean displacement gap} & \multicolumn{2}{c}{Majority of bases, \% layers} \\
\cmidrule(lr){2-3}\cmidrule(lr){4-5}\cmidrule(lr){6-7}
Model & unpaired & matched & unpaired & matched & contraction & displacement \\
\midrule
Pythia 2.8B  & 2.28 & 2.36 & 3.56 & 6.85 & 100 & 100 \\
Llama 3.2 3B & 1.81 & 1.75 & 0.74 & 2.02 & 97 & 100 \\
Qwen 2.5 3B  & 1.84 & 1.71 & 4.83 & 14.12 & 87 & 100 \\
Gemma 4 E4B  & 2.60 & 2.07 & 2.72 & 3.24 & 93 & 100 \\
\bottomrule
\end{tabular}}
\end{table}

Averaged over layers, the number of bases (of seven) in which refinement moves farther than its matched control is $5.6$ (Pythia), $5.6$ (Llama), $6.1$ (Qwen) and $6.6$ (Gemma).

\section{Quality-Controlled Rerun of the Refinement Analysis}
\label{app:qc}

Two mechanical filters, re-applicable to the released data, define the quality-controlled subset. \emph{Explicit cues}: every sentence containing ``More narrowly'', ``More precisely'', ``Specifically'', ``Strictly'', ``Technically'' or ``In precise terms'' ($243$ of $3{,}840$, $6.3\%$, all in template family T2). \emph{Malformed sentences}: every sentence ending in a dangling function word, e.g.\ ``The role of a form is.'' ($82$ sentences, $2.1\%$). Together, $314$ sentences ($8.2\%$) are removed ($11$ sentences fall under both rules).

Filtering creates a sample-size confound that must be removed before comparing. It drops $218$ of $240$ chain-level cells below $16$ sentences, and both spread measures are covariance quantities: an $n$-point centred cloud has $n-1$ non-zero eigenvalues, so log-volume sums a different number of terms as $n$ changes, and participation ratio is bounded above by $n-1$. Rerunning naively on the filtered data lowers the fraction of layers on which the refinement gap exceeds control from $98\%$ to $66\%$; this drop is arithmetic. Trimming every cell to a constant $n=12$ restores it to $97\%$, against $98\%$ for an unfiltered control trimmed to the same $n$, so we report the filtered analysis at matched sample size alongside that control.

At matched $n$ the effect is unchanged or slightly stronger in every model, each filtered estimate falls inside the $95\%$ band of $20$ random matched-$n$ draws, and per-layer sign agreement with the unfiltered curve is $97$--$100\%$. The three qualitative claims of Figure~\ref{fig:image5} are reproduced: refinement below control at $97\%$ of layers, monotone ordering by level at $100\%$, and $S_4$ contraction at $100\%$. Displacement agrees in sign on every layer in all four models; participation ratio stays directionally consistent at attenuated magnitude. Deleting any single template family (T1, T2, T3 or T4) preserves the log-volume contraction gap in all four models. Template family T1 contains no cue and no malformed sentences at all; restricted to T1 alone, the displacement gap stays positive in all four models ($+1.38$, $+0.34$, $+2.26$, $+2.06$).

\section{Euclidean Distance after Deflation and the Readout Split}
\label{app:euclid}

Under mean pooling, deflated Euclidean distance moves progressively toward B, the Mahalanobis ordering (Figure~\ref{fig:image3}). Table~\ref{tab:eucl_sweep} shows that the share of layers satisfying B grows with every additional removed component in all four models, reaching $81$--$100\%$ at top-$10$ deflation; at the lower deflation levels used in Table~\ref{tab:orderings}, Llama and Qwen have not yet reached it. Cosine similarity on the same deflated states satisfies B at no layer, at any $k$, in any model. Euclidean distance applies no inverse-variance weighting, but top-$k$ deflation is itself covariance-derived---a hard spectral filter that removes the top covariance directions---so the convergence shows that the ordering is not specific to inverse-variance weighting, not that it is independent of covariance-based preprocessing.

\begin{table}[H]
\centering
\small
\caption{Deflated Euclidean distance, mean pooling: percentage of layers satisfying B, with the percentage satisfying B$^{+}$ in parentheses, by deflation level $k$.}
\label{tab:eucl_sweep}
\begin{tabular}{lcccccc}
\toprule
Model & $k{=}5$ & $k{=}6$ & $k{=}7$ & $k{=}8$ & $k{=}9$ & $k{=}10$ \\
\midrule
Pythia 2.8B  & 0 (0)  & 30 (30) & 82 (79)  & 97 (88)  & 97 (91)   & 100 (91) \\
Llama 3.2 3B & 0 (0)  & 38 (34) & 100 (97) & 100 (97) & 100 (100) & 100 (100) \\
Qwen 2.5 3B  & 0 (0)  & 3 (3)   & 22 (16)  & 51 (41)  & 62 (54)   & 100 (95) \\
Gemma 4 E4B  & 16 (2) & 33 (19) & 44 (37)  & 67 (60)  & 81 (72)   & 81 (72) \\
\bottomrule
\end{tabular}
\end{table}

Under last-token readout, deflated Euclidean distance follows a different path. At top-$10$ deflation it satisfies B at only $12$--$27\%$ of layers in Llama, Qwen and Gemma, and strict R at $3$--$19\%$: unlike under mean pooling, it does not move toward the Mahalanobis ordering, and it does not separate the relatedness groups cleanly either. Pythia is the exception, with B at $94\%$ of layers. One possible account, which we have not tested directly, is the following. Mean pooling averages across token positions, which may suppress realization-specific variance and expose the shared covariance structure that the Mahalanobis metric formalizes, whereas last-token states retain the unaveraged representation at a single position, including realization-specific detail spread across many directions that deflation does not remove. Consistent with this account, the rank sweep (Appendix~\ref{app:rank}) shows that $r=128$ suffices to stabilize the Mahalanobis ordering under mean pooling, whereas last-token readout requires larger retained ranks.

\section{Full Refinement Geometry Tables}

The main text reports the refinement results in compressed form. For verification, the tables below summarize the cross-model contrast directly. Displacement is the more immediate effect, but contraction is the stronger neighborhood-level signature. Participation ratio is most informative where it diverges from the sentence-level contraction story and reveals a different token-local geometry.

\subsection{Displacement}

\begin{table}[H]
\centering
\small
\caption{Refinement-minus-control displacement for the $S_0 \rightarrow S_4$ contrast. Positive values mean refinement moves farther than control.}
\label{tab:disp_summary}
\resizebox{\textwidth}{!}{%
\begin{tabular}{llcrrrr}
\toprule
Model & Readout & Metric & Mean gap & AUC & Positive layers (\%) & Strict CI-separated (\%) \\
\midrule
Pythia 2.8B & Mean & Euclidean & 3.56 & 117.63 & 100 & 33 \\
Pythia 2.8B & Mean & Mahalanobis & 2.37 & 78.28 & 100 & 100 \\
Llama 3.2 3B & Mean & Euclidean & 0.74 & 21.58 & 100 & 21 \\
Llama 3.2 3B & Mean & Mahalanobis & 2.47 & 71.66 & 100 & 100 \\
Qwen 2.5 3B & Mean & Euclidean & 4.83 & 178.86 & 100 & 24 \\
Qwen 2.5 3B & Mean & Mahalanobis & 2.58 & 95.56 & 100 & 100 \\
Gemma 4 E4B & Mean & Euclidean & 2.72 & 116.83 & 100 & 93 \\
Gemma 4 E4B & Mean & Mahalanobis & 2.00 & 85.84 & 100 & 98 \\
Pythia 2.8B & Last-token & Euclidean & 0.63 & 20.90 & 94 & 33 \\
Pythia 2.8B & Last-token & Mahalanobis & 1.88 & 62.15 & 97 & 97 \\
Llama 3.2 3B & Last-token & Euclidean & 2.04 & 59.06 & 97 & 93 \\
Llama 3.2 3B & Last-token & Mahalanobis & 1.81 & 52.35 & 97 & 76 \\
Qwen 2.5 3B & Last-token & Euclidean & 6.37 & 235.65 & 97 & 92 \\
Qwen 2.5 3B & Last-token & Mahalanobis & 1.67 & 61.68 & 97 & 97 \\
Gemma 4 E4B & Last-token & Euclidean & 4.75 & 204.21 & 98 & 93 \\
Gemma 4 E4B & Last-token & Mahalanobis & 1.24 & 53.16 & 98 & 86 \\
\bottomrule
\end{tabular}
}
\end{table}

\subsection{Contraction}

\begin{table}[H]
\centering
\small
\caption{Refinement-minus-control log-volume summaries for span-4 contrasts. Positive values mean the control branch occupies the larger neighborhood, i.e., refinement is more contractive.}
\label{tab:lv_summary}
\resizebox{\textwidth}{!}{%
\begin{tabular}{lllrrrr}
\toprule
Model & Readout & Condition & Mean gap & AUC & Positive layers (\%) & Strict CI-separated (\%) \\
\midrule
Pythia 2.8B & Mean & Raw & 2.28 & 75.08 & 97 & 94 \\
Pythia 2.8B & Mean & Top-5 deflated & 3.37 & 111.22 & 100 & 97 \\
Llama 3.2 3B & Mean & Raw & 1.81 & 52.52 & 100 & 66 \\
Llama 3.2 3B & Mean & Top-5 deflated & 2.58 & 74.95 & 100 & 97 \\
Qwen 2.5 3B & Mean & Raw & 1.84 & 68.20 & 100 & 62 \\
Qwen 2.5 3B & Mean & Top-5 deflated & 2.50 & 92.51 & 100 & 86 \\
Gemma 4 E4B & Mean & Raw & 2.60 & 111.89 & 100 & 91 \\
Gemma 4 E4B & Mean & Top-5 deflated & 2.66 & 114.21 & 100 & 81 \\
Pythia 2.8B & Last-token & Raw & 1.45 & 48.00 & 94 & 33 \\
Pythia 2.8B & Last-token & Top-5 deflated & 1.76 & 57.92 & 97 & 64 \\
Llama 3.2 3B & Last-token & Raw & 2.62 & 75.99 & 93 & 83 \\
Llama 3.2 3B & Last-token & Top-5 deflated & 3.72 & 107.77 & 97 & 93 \\
Qwen 2.5 3B & Last-token & Raw & 1.22 & 45.15 & 78 & 35 \\
Qwen 2.5 3B & Last-token & Top-5 deflated & 2.50 & 92.54 & 95 & 73 \\
Gemma 4 E4B & Last-token & Raw & 1.50 & 64.63 & 88 & 26 \\
Gemma 4 E4B & Last-token & Top-5 deflated & 2.71 & 116.70 & 95 & 53 \\
\bottomrule
\end{tabular}
}
\end{table}

\subsection{Participation Ratio}

\begin{table}[H]
\centering
\small
\caption{Refinement-minus-control raw participation-ratio summaries for span-4 contrasts. Negative values indicate lower effective dimensionality under refinement than under control.}
\label{tab:pr_summary}
\resizebox{\textwidth}{!}{%
\begin{tabular}{llrrrrrr}
\toprule
Model & Readout & Mean gap & AUC & Positive (\%) & Negative (\%) & Strict positive (\%) & Strict negative (\%) \\
\midrule
Pythia 2.8B & Mean & -0.70 & -23.25 & 55 & 45 & 42 & 33 \\
Llama 3.2 3B & Mean & -0.26 & -7.61 & 66 & 34 & 0 & 21 \\
Qwen 2.5 3B & Mean & -0.37 & -13.56 & 0 & 100 & 0 & 22 \\
Gemma 4 E4B & Mean & -0.68 & -29.38 & 0 & 100 & 0 & 67 \\
Pythia 2.8B & Last-token & -0.08 & -2.72 & 61 & 36 & 0 & 15 \\
Llama 3.2 3B & Last-token & -2.63 & -76.13 & 0 & 97 & 0 & 97 \\
Qwen 2.5 3B & Last-token & -2.00 & -73.95 & 0 & 97 & 0 & 89 \\
Gemma 4 E4B & Last-token & -1.61 & -69.18 & 2 & 95 & 0 & 70 \\
\bottomrule
\end{tabular}
}
\end{table}

\section{Adversarial Controls}
\label{sec:adv_appendix}

The adversarial arms address a specific concern: that the refinement--control contrast could be an artifact of how the control arm is defined, rather than a signature of semantic narrowing. The control arm adds attributive content that stays within the base concept's typical distribution; a sceptic could argue that any arm built under a different generative rule would produce the same divergence from refinement, making the contrast uninformative about meaning. Near-synonym stacking and scope hedging provide two independent stress tests: they each follow a different rule from control but implement neither the semantic boundary of refinement nor the attributive accumulation of control. If the contrast between refinement and control were merely definitional, these arms should track control closely. The figures below show that they do not do so uniformly, which is consistent with the refinement effect reflecting something specific to semantic narrowing; given the arm sizes, this is supporting rather than primary evidence.

\begin{figure}[H]
\centering
\includegraphics[width=\textwidth]{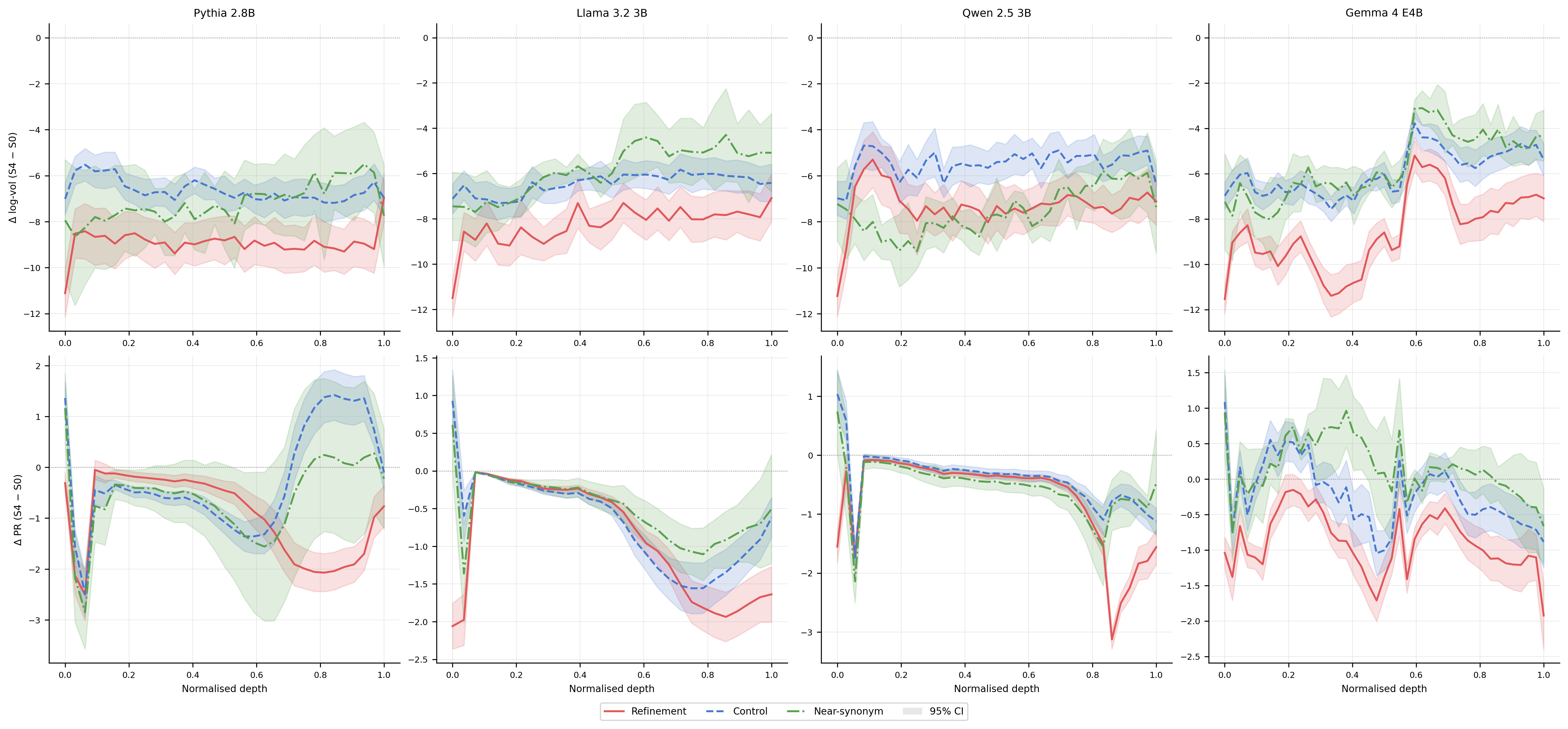}
\caption{Gap profiles $\Delta\mathrm{metric}(S_4{-}S_0)$ for refinement (solid), control (dashed), and near-synonym (dash-dot), with 95\% CI bands shaded. Top row: log-volume; bottom row: participation ratio. Columns are models. Near-synonym tracks close to refinement in log-volume for Pythia and Qwen, while Llama and Gemma show a shallower contraction. In participation-ratio terms the arm is usually near zero or positive, contrasting with refinement which is always negative.}
\label{fig:adv_near_synonym}
\end{figure}

\begin{figure}[H]
\centering
\includegraphics[width=\textwidth]{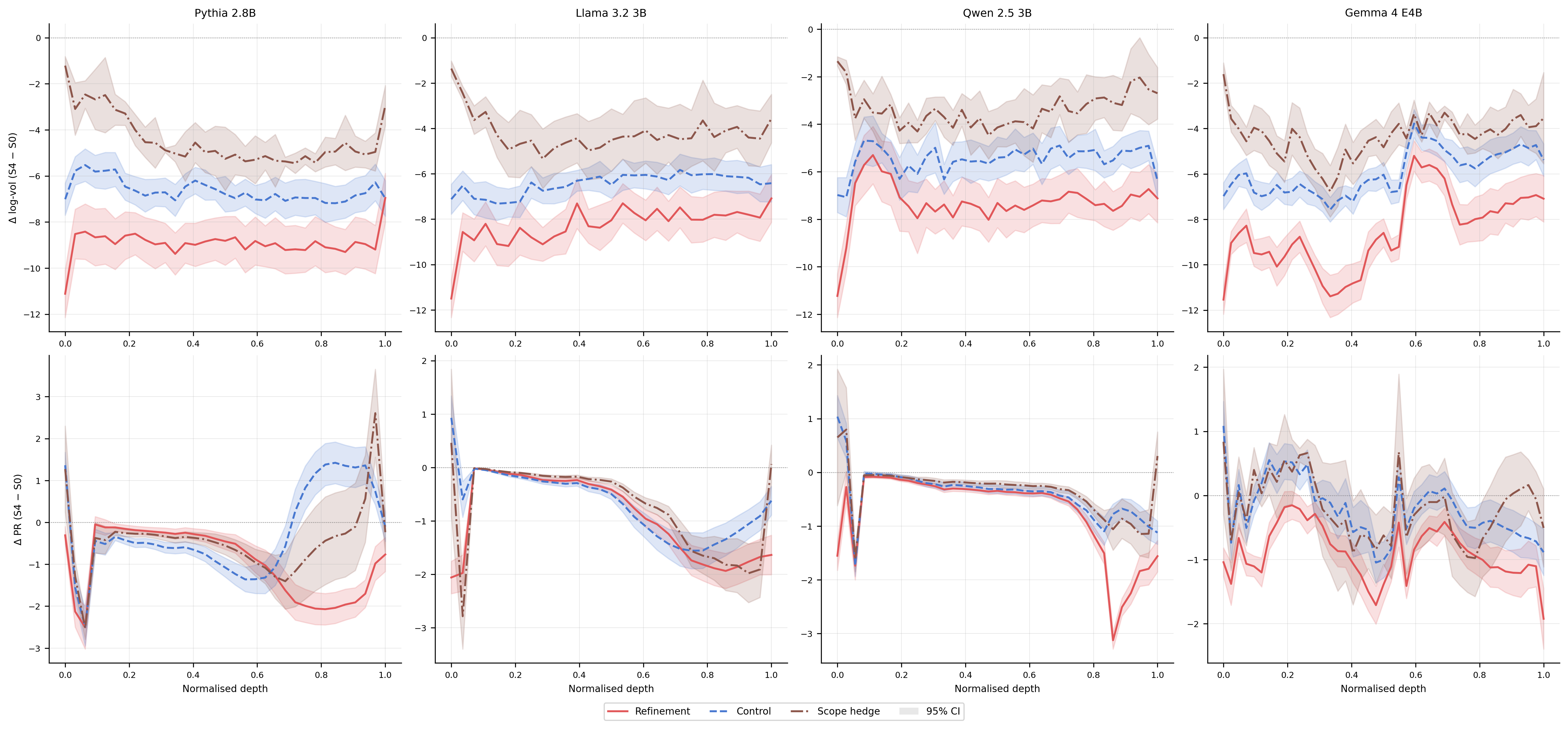}
\caption{Same layout as Figure~\ref{fig:adv_near_synonym} for scope-hedge. The separation from refinement is consistent across all four models and both metrics: scope-hedge produces a substantially smaller log-volume contraction and stays near or above zero in participation ratio throughout the network depth, whereas refinement is persistently more negative in both.}
\label{fig:adv_scope_hedge}
\end{figure}

Figures~\ref{fig:adv_near_synonym} and~\ref{fig:adv_scope_hedge} show the full depth profiles for log-volume (top row) and participation ratio (bottom row) across all four models, separately for each adversarial arm. Each adversarial arm has only $n{=}4$ chains versus $n{=}20$ for refinement and control; the resulting 95\% CI bands are visibly wider for the adversarial arm in every panel, which bounds how confident any layer-specific inference can be.

Log-volume (top rows) is the more informative metric. Scope-hedge is consistently the shallowest arm in all four models, indicating that hedging commitment strength does not reproduce the directed volume compression of refinement. Near-synonym is less clear-cut: it overlaps substantially with refinement in Pythia and Qwen, but shows shallower contraction in Llama and Gemma; the wide CI bands ($n{=}4$) prevent firm layer-specific conclusions. In participation-ratio terms (bottom rows) the contrast is sharper for both arms: refinement stays negative across the full depth profile, while the adversarial controls are usually near zero or positive, with overlap only at very early layers.

\section{Token Length by Level in the Refinement Dataset}
\label{sec:token_length}

A potential confound in the refinement probe is that geometric displacement or neighborhood contraction could be driven by sentence length rather than semantic content. Table~\ref{tab:token_counts} reports the mean number of tokens per sentence at each level $S_0$--$S_4$, computed with each model's own tokenizer. Length growth across levels is minimal: the largest increase from $S_0$ to $S_4$ is $0.8$ tokens (Pythia, near-synonym), and the majority of arms show increases of $0.3$--$0.6$ tokens. Length therefore cannot account for the observed geometry.

\begin{table}[H]
\centering
\small
\caption{Mean number of tokens per sentence at each refinement level, by arm and model. Token counts use each model's own tokenizer. Growth from $S_0$ to $S_4$ is $0.2$--$0.8$ tokens across all conditions.}
\label{tab:token_counts}
\begin{tabular}{llrrrrr}
\toprule
Arm & Model & $S_0$ & $S_1$ & $S_2$ & $S_3$ & $S_4$ \\
\midrule
\multirow{4}{*}{Refinement}
 & Pythia 2.8B   & 10.5 & 10.6 & 10.8 & 10.9 & 11.1 \\
 & Llama 3.2 3B  & 11.4 & 11.4 & 11.5 & 11.5 & 11.8 \\
 & Qwen 2.5 3B   & 10.4 & 10.4 & 10.5 & 10.5 & 10.8 \\
 & Gemma 4 E4B   & 11.4 & 11.4 & 11.4 & 11.5 & 11.6 \\
\midrule
\multirow{4}{*}{Control}
 & Pythia 2.8B   & 11.5 & 11.5 & 11.7 & 11.8 & 11.9 \\
 & Llama 3.2 3B  & 12.3 & 12.4 & 12.5 & 12.6 & 12.7 \\
 & Qwen 2.5 3B   & 11.3 & 11.4 & 11.5 & 11.6 & 11.7 \\
 & Gemma 4 E4B   & 12.3 & 12.3 & 12.4 & 12.4 & 12.5 \\
\midrule
\multirow{4}{*}{Near-synonym}
 & Pythia 2.8B   & 11.4 & 11.5 & 11.5 & 11.7 & 12.2 \\
 & Llama 3.2 3B  & 12.3 & 12.4 & 12.3 & 12.3 & 12.6 \\
 & Qwen 2.5 3B   & 11.3 & 11.4 & 11.3 & 11.3 & 11.6 \\
 & Gemma 4 E4B   & 12.2 & 12.3 & 12.3 & 12.3 & 12.5 \\
\midrule
\multirow{4}{*}{Scope-hedge}
 & Pythia 2.8B   & 10.9 & 11.0 & 11.0 & 11.2 & 11.2 \\
 & Llama 3.2 3B  & 11.9 & 11.9 & 12.0 & 12.1 & 12.1 \\
 & Qwen 2.5 3B   & 10.9 & 10.9 & 11.0 & 11.1 & 11.1 \\
 & Gemma 4 E4B   & 11.9 & 11.9 & 12.0 & 12.1 & 12.1 \\
\bottomrule
\end{tabular}
\end{table}

\end{document}